\documentclass{article}

\usepackage{arxiv}

\usepackage[utf8]{inputenc} 
\usepackage[T1]{fontenc}    
\usepackage[pagebackref=true,breaklinks=true,colorlinks,bookmarks=false]{hyperref} 
\usepackage{url}            
\usepackage{booktabs}       
\usepackage{amsfonts}       
\usepackage{nicefrac}       
\usepackage{microtype}      
\usepackage{lipsum}		
\usepackage{graphicx}
\usepackage{natbib}
\usepackage{doi}

\title{Sound and Complete Neural Network Repair\\ with Minimality and Locality Guarantees}

\author{{\hspace{1mm}Feisi Fu}\\
	Division of Systems Engineering\\
	Boston University\\
	Boston, MA 02215 \\
	\texttt{fufeisi@bu.edu} \\
	\And
	{\hspace{1mm}Wenchao Li} \\
	Department of Electrical and Computer Engineering\\
	Boston University\\
	Boston, MA 02215 \\
	\texttt{wenchao@bu.edu} \\
}

\date{}

\renewcommand{\undertitle}{}
\renewcommand{\shorttitle}{\toolname}

\usepackage{amsfonts} 
\usepackage{amsthm} 
\usepackage{amssymb}
\usepackage{amsmath}
\usepackage[utf8]{inputenc}
\usepackage{graphicx}
\usepackage{subcaption}
\usepackage[all]{xy} 
\usepackage{tkz-euclide}
\tikzset{elegant/.style={smooth,thick,samples=50,cyan}}
\tikzset{eaxis/.style={->,>=stealth}}
\usepackage{multirow}

\usepackage{algorithm}
\usepackage{algorithmic}

\usepackage{blkarray} 

\usepackage{enumitem} 

\newtheorem{definition}{Definition}

\newtheorem{theorem}{Theorem}
\newtheorem{example}{Example}
\newtheorem{lemma}{Lemma}
\newtheorem{specification}{Specification}
\newtheorem{corollary}{Corollary}

\usepackage{xspace}
\newcommand{\patcharea}{\mathcal{A}\xspace}
\newcommand{\toolname}{\texttt{REASSURE}\xspace}
\newcommand{\buggypoint}{\widetilde{x}}

\usepackage{xcolor}

\newcommand{\commentout}[1]{}

\newtheorem{innercustomgeneric}{\customgenericname}
\providecommand{\customgenericname}{}
\newcommand{\newcustomtheorem}[2]{%
  \newenvironment{#1}[1]
  {%
   \renewcommand\customgenericname{#2}%
   \renewcommand\theinnercustomgeneric{##1}%
   \innercustomgeneric
  }
  {\endinnercustomgeneric}
}

\newcustomtheorem{customthm}{Theorem}

\usepackage{lineno} 

\begin{document}
\maketitle

\begin{abstract}
We present a novel methodology for repairing neural networks that use ReLU activation functions. Unlike existing methods that rely on modifying the weights of a neural network which can induce a global change in the function space, our approach applies only a localized change in the function space while still guaranteeing the removal of the buggy behavior. By leveraging the piecewise linear nature of ReLU networks, our approach can efficiently construct a patch network tailored to the linear region where the buggy input resides, which when combined with the original network, provably corrects the behavior on the buggy input. Our method is both sound and complete -- the repaired network is guaranteed to fix the buggy input, and a patch is guaranteed to be found for any buggy input. Moreover, our approach preserves the continuous piecewise linear nature of ReLU networks, automatically generalizes the repair to all the points including other undetected buggy inputs inside the repair region, is minimal in terms of changes in the function space, and guarantees that outputs on inputs away from the repair region are unaltered. On several benchmarks, we show that our approach significantly outperforms existing methods in terms of locality and limiting negative side effects. Our code is available on GitHub: \url{https://github.com/BU-DEPEND-Lab/REASSURE}.
\end{abstract}

\section{Introduction}
   
Deep neural networks (DNNs) have demonstrated impressive performances on a wide variety of applications ranging from transportation~\cite{nvidia-dave} to health care~\cite{nn-healthcare}. However, DNNs are not perfect. 
In many cases, especially when the DNNs are used in safety-critical contexts, it is important to correct erroneous outputs of a DNN as they are discovered after training. 
For instance, a neural network in charge of giving control advisories to the pilots in an aircraft collision avoidance system, such as the ACAS Xu network from \cite{julian2019deep}, may produce an incorrect advisory for certain situations and cause the aircraft to turn towards the incoming aircraft, thereby jeopardizing the safety of both airplanes.
In this paper, we consider the problem of \textit{neural network repair}, i.e. given a trained neural network and a set of buggy inputs (inputs on which the neural network produces incorrect predictions)
, repair the network so that the resulting network on those buggy inputs behave according to some given correctness specification. 
Ideally, the changes to the neural network function should be small so that the outputs on other inputs are either unchanged or altered in a small way. 
Existing efforts on neural network repair roughly fall into the following three categories.

    \quad1. \textit{Retraining/fine-tuning.} 
    The idea is to retrain or fine-tune the network with the newly identified buggy inputs and the corresponding corrected outputs. Methods include counterexample-guided data augmentation~\cite{dreossi2018counterexample,ren2020few},  
    editable training~\cite{sinitsin2020editable} and
    training input selection~\cite{ma2018mode}.
    One major weakness of these approaches is the lack of formal guarantees -- at the end of retraining/fine-tuning, there is no guarantee that the given buggy inputs are fixed and no new bugs are introduced. 
    In addition, retraining can be very expensive and requires access to the original training data which is impractical in cases where the neural network is obtained from a third party or the training data is private. Fine-tuning, on the other hand, often faces the issue of catastrophic forgetting~\cite{catastrophic-forgetting}. 
    
    \quad2. \textit{Direct weight modification.} 
    These approaches directly manipulate the weights in a neural network to fix the buggy inputs.
    The repair problem is typically cast into an optimization problem or a verification problem. For example, \cite{dong2020towards} proposes to minimize a loss defined based on the buggy inputs. \cite{goldberger2020minimal} uses an SMT solver to identify minimal weight changes to the output layer of the network so that the undesirable behaviors are removed.
    In general, the optimization-based approach cannot guarantee removal of the buggy behaviors, and the verification-based approach does not scale beyond networks of a few hundred neurons.
    In addition, both approaches suffer from substantial accuracy drops on normal inputs since \textit{weight changes may be a poor proxy for changes in the function space}. 
    
    
    \quad3. \textit{Architecture extension.} 
    The third category of approaches extends the given NN architecture, such as by introducing more weight parameters, to facilitate more efficient repairs. The so-called Decoupled DNN architecture~\cite{sotoudeh2021provable} is the only work we know that falls into this category. Their idea is to decouple the activations of the network from values of the network by augmenting the original network. Their construction allows the formulation of any single-layer repair as an linear programming (LP) problem. The decoupling, however, causes the repaired network to become discontinuous (in the functional sense). In addition, it still cannot isolate the output change to a single buggy input from the rest of the inputs.
    
    \begin{figure}
        \centering
        \includegraphics[width=0.75\textwidth]{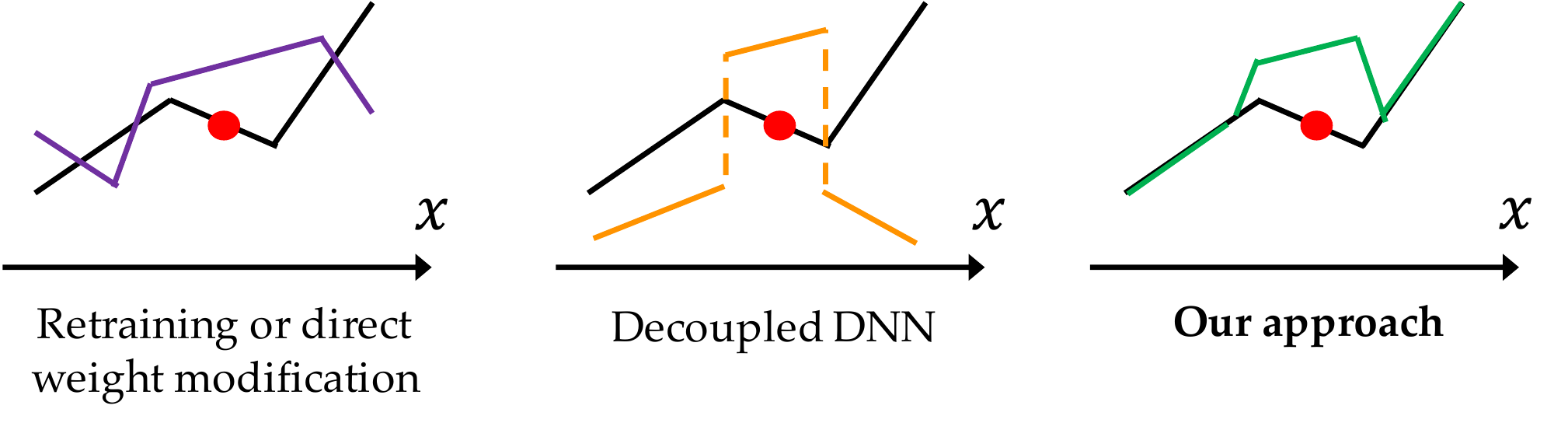}
        \caption{Comparison of different approaches to the neural network repair problem. The black lines represent the original neural network function. The red dot represents the buggy input. The colored lines represent the functions after the repairs are done.}
        \label{fig: repair}
        \vspace{-4mm}
    \end{figure}
    
    In addition to the aforementioned limitations, a common weakness that is shared amongst these methods is that the induced changes, as a result of either retraining or direct weight modification, are \textit{global}.
    This means that a correct behavior on another input, regardless of how far it is away from the buggy input, may not be preserved by the repair. 
    Worse still, the repair on a new buggy input can end up invalidating the repair on a previous buggy input. 
    The fundamental issue here is that limiting the changes to a few weights or a single layer only poses a structural constraint (often for ease of computation); it does not limit the changes on the input-output mapping of the neural network.
    It is known that even a single weight change can have a global effect on the output of a neural network.

    In this paper, we propose \toolname, 
    a novel methodology for neural network repair with locality, minimality, soundness and completeness guarantees.
    Our methodology targets continuous piecewise linear (CPWL) neural networks, specifically those that use the ReLU activation functions. 
    The key idea of our approach is to \textit{leverage the CPWL property of ReLU networks to synthesize a patch network tailored to the linear region where the buggy input resides, which when combined with the original network, provably corrects the behavior on the buggy input}.
    Our approach is both sound and complete -- the repaired network is guaranteed to fix the buggy input, and a patch is guaranteed to be found for any buggy input. Moreover, our approach preserves the CPWL nature of ReLU networks, automatically generalizes the repair to all the points including other undetected buggy inputs inside the repair region, is minimal in terms of changes in the function space, and guarantees that outputs on inputs away from the repair region are unaltered. 
    Figure~\ref{fig: repair} provides an illustrative comparison of our approach with other methods. 
    Table~\ref{tab: requirements} compares our approach with representative related works in terms of theoretical guarantees. 
    We summarize our contributions below.
    \begin{itemize}
        \item We present \toolname, the first sound and complete repair methodology for ReLU networks with strong theoretical guarantees. 
        \item Our technique synthesizes a patch network, which when combined with the original neural network, provably corrects the behavior on the buggy input. This approach is a significant departure from existing methods that rely on retraining or direct weight manipulation.
        \item Our technique is both effective and efficient -- across a set of benchmarks, \toolname can efficiently correct a set of buggy inputs or buggy areas with little or no change to the accuracy and overall functionality of the network. 
\end{itemize}

\section{Background}
    \begin{table*}
        \centering
        \scalebox{0.9}{
        \begin{tabular}{c|ccccc}
        & \toolname & Retrain & MDNN & Editable Fine-Tuning & PRDNN \\
        \hline
        Preservation of CPWL & Yes & Yes & Yes & Yes & No\\
        Soundness & Yes & No & Yes & No & Yes \\
        Completeness & Yes & No & No & No & No \\
        Area Repair & Yes & No & No & No & Yes \\
        Minimal Change & Yes (Function Space) & No & Yes (Weight Space) & No & Yes (Weight Space) \\
        Localized Change & Yes & No & No & No & No \\
        Limited Side Effect & Yes & No & No & No & No
        \end{tabular}
        }
        \caption{Comparing \toolname with representative related works in terms of theoretical guarantees. 
        CPWL stands for continuous piecewise linearity. 
        Area repair means repairing all the (infinitely many) points inside an area. 
        Limited side effect means the repair can limit potential adverse effects on other inputs. 
        MDNN is the verification-based approach from \cite{goldberger2020minimal}. 
        PRDNN is the Decoupled DNN approach from \cite{sotoudeh2021provable}.
        \toolname is the only method that can provide all the guarantees.
        }
        \label{tab: requirements}
        \vspace{-2mm}
    \end{table*}
    \subsection{Deep Neural Networks}
        An $R$-layer feed-forward DNN $f = \kappa_R \circ \sigma \circ \kappa_{R-1} \circ... \circ \sigma \circ \kappa_1: X\to Y$ is a composition of linear functions $\kappa_r, r=1, 2, ..., R$ and activation function $\sigma$, where $X\subseteq \mathbb{R}^m$ is a bounded input domain and $Y\subseteq\mathbb{R}^n$ is the output domain.
        Weights and biases of linear function $\{\kappa_r\}_{r=1, 2,...,R}$ are parameters of the DNN.
        
        We call the first $R - 1$ layers hidden layers and the $R$-th layer the output layer. 
        We use $z^i_j$ to denote the $i$-th neuron (before activation) in the $j$-th hidden layer.
        
        In this paper, we focus on ReLU DNNs, i.e. DNNs that use only the ReLU activation functions. 
        It is known that an $\mathbb{R}^m \rightarrow \mathbb{R}$ function is representable by a ReLU DNN \textit{if and only if} it is a continuous piecewise linear (CPWL) function~\cite{arora2016relu}. 
        The ReLU function is defined as $\sigma(x) = \text{max}(x, 0)$. We say that $\sigma(x)$ is activated if $\sigma(x) = x$.
        
        
    \subsection{Linear Regions}
        A linear region $\patcharea$ is the set of inputs
        that correspond to the same activation pattern in a ReLU DNN $f$~\cite{serra2017linear-region}.
        Geometrically, this corresponds to a convex polytope, which is an intersection of half spaces, in the input space $X$ on which $f$ is linear. We use $f|_\patcharea$ to denote the part of $f$ on $\patcharea$. 
    
    \subsection{Correctness Specification}
    
        A correctness specification $\Phi = (\Phi_{in}, \Phi_{out})$ is a tuple of two polytopes, where 
        $\Phi_{in}$ is the union of some linear regions and $\Phi_{out}$ is a convex polytope. A DNN $f$ is said to meet a specification $\Phi = (\Phi_{in}, \Phi_{out})$, denoted as $f \models \Phi$, if and only if $\forall x \in \Phi_{in}, f(x)\in \Phi_{out}$.
        
    
        \begin{example}\label{exp: classification problem}
            For a classification problem, we can formally write the specification that ``the prediction of any point in an area $\patcharea$ is class $k$" as $\Phi = (\Phi_{in}, \Phi_{out})$, where $\Phi_{in} = \patcharea$ and $\Phi_{out} = \{y\in \mathbb{R}^n\:|\: y_k\geq y_i, \forall i\neq k\}$\footnote{ 
            Note that here $y$ is the output of the layer right before the softmax layer in a classification network.}.
        \end{example}
    
    
    
    \subsection{Problem Definition}
        In this paper, we consider the following two repair problems. 
        \begin{definition}[Area repair]\label{def: area repair}
            Given a correctness specification $\Phi = (\Phi_{in}, \Phi_{out})$ and a ReLU DNN $f \not\models \Phi$, the area repair problem is to find a modified ReLU DNN $\widehat{f}$ such that $\widehat{f} \models \Phi$.
        \end{definition}
        
        Note that we do not require $\widehat{f}$ to have the same structure or parameters as $f$ in this definition.
        If $\Phi_{in}$ contains a single (buggy) linear region, we refer to this as \textit{single-region repair}. If $\Phi_{in}$ contains multiple (buggy) linear regions, we refer to it as \textit{multi-region repair}.

        \begin{definition}[Point-wise repair]\label{def: point-wsie repair}
            Given a set of buggy inputs $\{\buggypoint_1,\ldots,\buggypoint_L\} \subset \Phi_{in}$ with their corresponding correct outputs $\{y_1,\ldots,y_L\}$ and a ReLU DNN $f$, the point-wise repair problem is to
            find a modified ReLU DNN $\widehat{f}$ such that $\forall i, \widehat{f}(\buggypoint_i) = y_i$.
        \end{definition}
    
        We call the minimal variants of area repair and point-wise repair \textit{minimal area repair} and \textit{minimal point-wise repair} respectively. Minimality here is defined with respect to the maximum distance between $f$ and $\widehat{f}$ over
        the whole input domain $X$.
        A point-wise repair can be generalized to an area repair through the following result. 
    
    \subsection{From Buggy Inputs to Buggy Linear Regions}\label{sec: point-wise}
        The linear region where an input $x$ resides can be computed as follows.

        \begin{lemma}\cite{lee2019towards}
            Consider a ReLU DNN $f$ and an input $x\in X$. For every neuron $z^i_j$, it induces a feasible set 
            \begin{eqnarray}\label{equ: buggy point to linregion}
                \patcharea^i_j(x) = 
                \begin{cases}
                    \{\bar{x}\in X|(\triangledown_x z^i_j)^T \bar{x} + z^i_j - (\triangledown_x z^i_j)^T x \geq 0\} \textit{ \qquad if $z^i_j\geq 0$}\\
                    \{\bar{x}\in X|(\triangledown_x z^i_j)^T \bar{x} + z^i_j - (\triangledown_x z^i_j)^T x \leq 0\} \textit{ \qquad if $z^i_j < 0$}
                \end{cases}
            \end{eqnarray}
            
            The set $\patcharea(x) = \cap_{i,j} \patcharea^i_j(x)$ is the linear region that includes $x$. Note that $\patcharea(x)$ is essentially the H-representation of the corresponding convex polytope.
        \end{lemma}
        
        

    \subsection{Repair Desiderata}
        We argue that an effective repair algorithm for ReLU DNN should satisfy the following criteria. 
        
        \quad\textbf{Preservation of CPWL}: 
        Given that the original network $f$ models a CPWL function, the repaired network $\widehat{f}$ should still model a CPWL function.
        
        \quad\textbf{Soundness}: 
        A sound repair should completely remove the known buggy behaviors, i.e. it is a solution to 
        the point-wise repair problem defined in Definition~\ref{def: point-wsie repair}. 
        
        \quad\textbf{Completeness}:
        Ideally, the algorithm should always be able find a repair for any given buggy input if it exists.
        
        \quad\textbf{Generalization}: 
        If there exists another buggy input $\widetilde{x}'$ in the neighborhood of $\widetilde{x}$ (e.g. the same linear region), then the repair should also fix it. 
        For example, suppose we have an $\widetilde{x}$ that violates a specification which requires the output to be within some range. It is almost guaranteed
        that there exists another (and infinitely many) $\widetilde{x}'$ in the same linear region that also violates the specification.
        
        \quad\textbf{Locality}: 
        We argue that a good repair should only induce a localized change to $f$ in the function space. For example, in the context of ReLU DNN, if a linear region $\mathcal{B}$ does not border the repair region $\patcharea$, i.e. 
        $\mathcal{B} \cap \patcharea = \emptyset$, 
        then $\widehat{f}|_\mathcal{B}(x) = f|_\mathcal{B}(x)$.
      
        \quad\textbf{Minimality}: 
        Some notion of distance between $f$ and $\widehat{f}$ such as $\max|f - \widehat{f}|$ should be minimized. Note that this is a significant departure from existing methods that focus on minimizing the change in weights which has no guarantee on the amount of change in the function space. 
        
        \quad\textbf{Limited side effect}: 
        Repairing a buggy point should not adversely affect points that were originally correct. 
        For example, repairing a buggy input $\widetilde{x}$
        in region $\patcharea$ should not change another region from correct to incorrect.
        Formally, for any linear region $\mathcal{C}$ who is a neighbor of $\patcharea$, i.e. $\mathcal{C} \cap \patcharea \neq \emptyset$, if $f|_\mathcal{C} \models \Phi$, then $\widehat{f}|_\mathcal{C} \models \Phi$. 
        
        \quad\textbf{Efficiency}: The repair algorithm should terminate in polynomial time with respect to the size of the neural network and the number of buggy inputs.
    
\section{Our Approach}
    We will first describe our approach to \textit{single-region repair} and then present our approach to \textit{multi-region repair} which builds on results obtained from the single-region case.
    
    Given a linear region $\patcharea$, our overarching approach is to synthesize a \textit{patch network} $h_\patcharea$ such that $\widehat{f} = f + h_\patcharea$ and $\widehat{f} \models \Phi$.
    The patch network $h_\patcharea$ is a combination of two sub-networks: a support network $g_\patcharea$ which behaves like a characteristic function to ensure that $h_\patcharea$ is almost only active on $\patcharea$, and an affine patch function network $p_\patcharea(x) = \boldsymbol{c}x+d$ such that $(f + p_\patcharea) \models \Phi$ on $\patcharea$.
    \subsection{Running Example}
    We use the following example to illustrate our idea.
    
\begin{figure}
        \centering
        \includegraphics[width=0.4\textwidth]{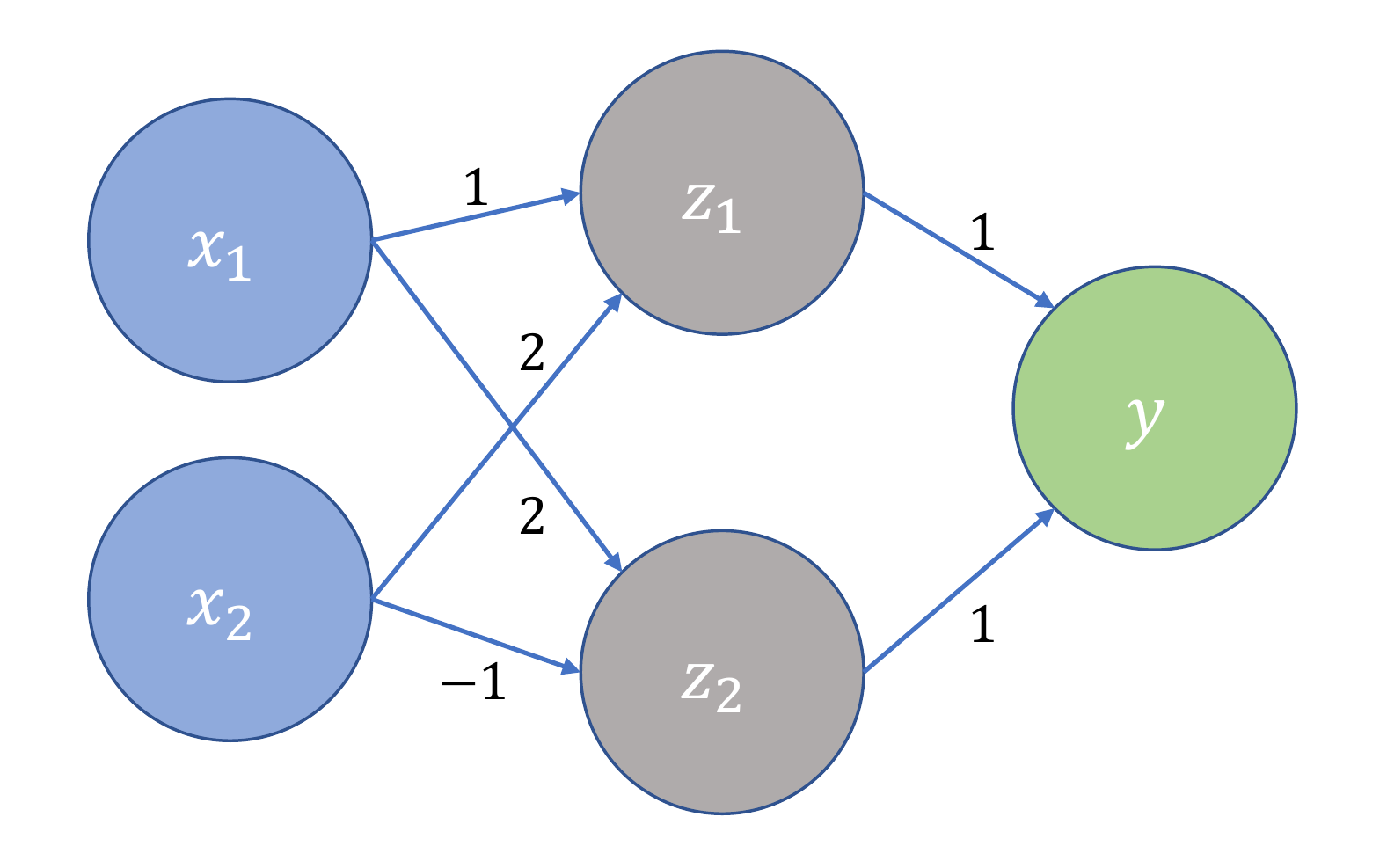}
        \includegraphics[width=0.45\textwidth]{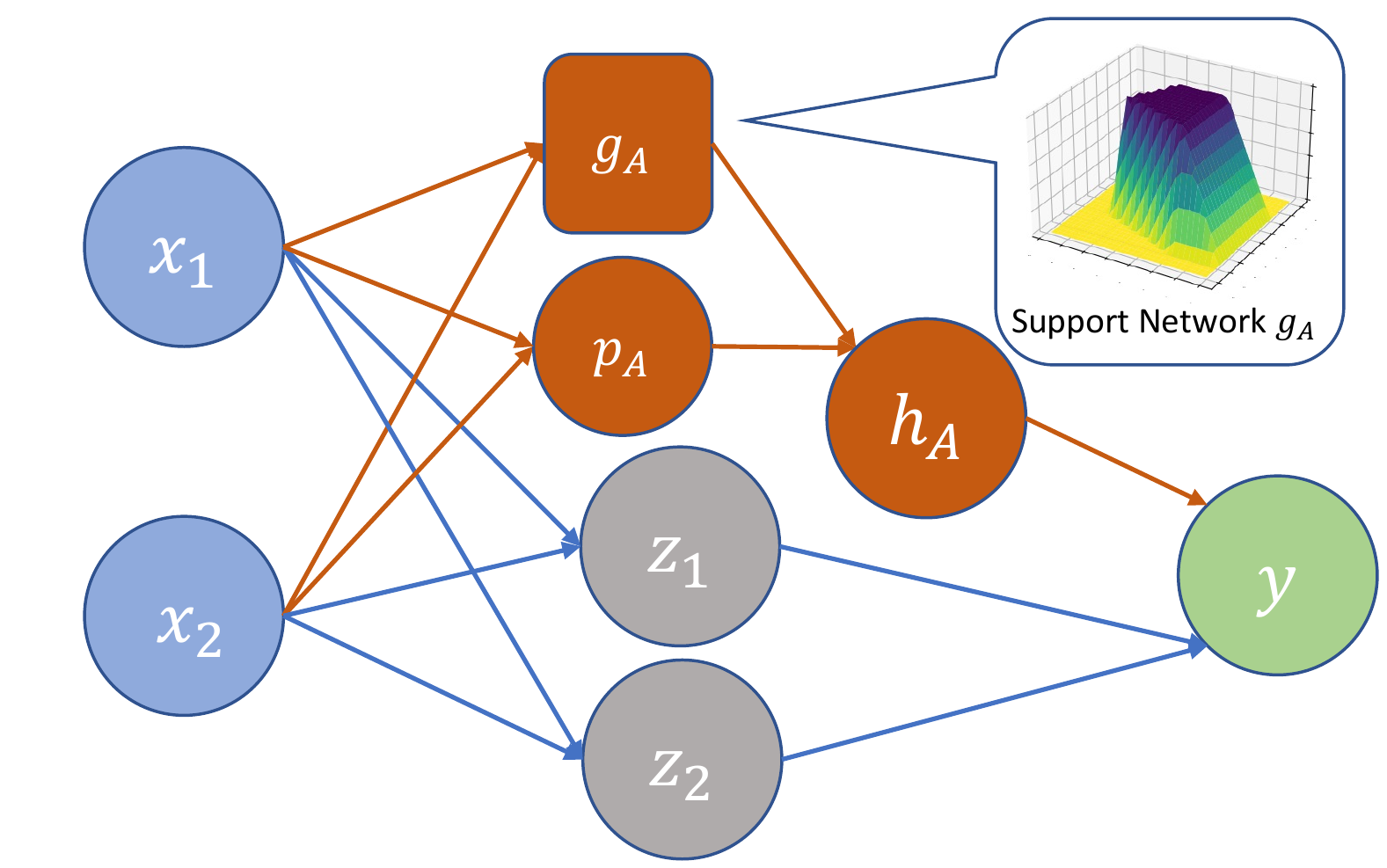}
        \caption{Left: the target DNN with buggy inputs. 
        Right: the \toolname-repaired DNN with the patch network shown in red. Support network $g_\patcharea$ is for approximating the characteristic function on $\patcharea$; Affine patch function $p_\patcharea$ ensures the satisfaction of $\Phi$ on $\patcharea$; The design of the patch network $h_\patcharea$ ensures locality for the final patch.
        }\label{fig: RunEx}
        \vspace{-2mm}
    \end{figure}

    \begin{example}
        Consider repairing the ReLU DNN $f$ in Figure~\ref{fig: RunEx} 
        according to the correctness specification $\Phi: \forall x \in [0, 1]^2$, $y\in [0, 2]$. 
        The DNN consists of a single hidden layer with two neurons $z_1$ and $z_2$, where $y = \sigma(z_1) + \sigma(z_2)$, $z_1 = x_1+2x_2-1$ and $z_2 = 2x_1-x_2$.
    \end{example}
    
    
    The only linear region that violates our specification is $\patcharea = \{x\:|\:1\geq x_1, 1\geq x_2, x_1+2x_2-1\geq 0, 2x_1-x_2\geq 0\}$. ( $\widetilde{x} = (0.9, 0.9)\in [0, 1]^2$ but $f(\widetilde{x}) = 2.6 \notin [0, 2]$)
    
    The network $f(x)$ on the linear region $\patcharea$ is the affine function $f|_\patcharea(x) = 3x_1+x_2-1$. 
    Our algorithm first sets up an affine function $p_\patcharea(x)$ that minimally repairs $f$ on $\patcharea$, such that $\forall x \in \patcharea, f(x)+p_\patcharea(x)\in [0, 2]$. 
    Later in the paper, we will show $p_\patcharea(x)$ can be found by solving a LP problem. 
    The resulting patch function is $p_\patcharea(x) = -\frac{1}{2}x_1-\frac{1}{2}x_2$. 
    
    However, directly apply $f(x) + p_\patcharea(x)$ as the patch network will have side effects on areas outside of $\patcharea$. 
    Our strategy is to combine $p_\patcharea(x)$ with a support network $g_\patcharea(x)$ 
    which outputs $1$ on $\patcharea$ and drops to $0$ quickly outside of $\patcharea$.
    The final repaired network is $f(x) + \sigma(p_\patcharea(x)+g_\patcharea(x,10)-1) - \sigma(-p_\patcharea(x)+g_\patcharea(x,10)-1)$.  This structure makes $p_\patcharea$ almost only active on $\patcharea$ and achieve a localized repair.
    Observe that this is still a ReLU DNN.

    \subsection{Support Networks}
        
        
        Support networks are neural networks with a special structure that can approximate the characteristic function of a convex polytope. They are keys to ensuring localized repairs in our algorithm.
        
        Assume that the linear region we need to repair is $\patcharea = \{x\:|\:a_ix\leq b_i, i\in I\}$, 
        where $|I|$ is the number of linear inequalities. 
        The support network of $\patcharea$ is defined as: 
        \begin{eqnarray}\label{eqn: support neural network}
            g_\patcharea(x, \gamma) = \sigma(\sum_{i\in I} g(b_i-a_ix, \gamma)-|I|+1)
        \end{eqnarray}
        where $g(x, \gamma) = \sigma(\gamma x+1) - \sigma(\gamma x)$ and $\gamma \in \mathbb{R}$ is a parameter of our algorithm that controls the slope of support networks.
        
        \noindent
        \textbf{Remark}: For any $x \in \patcharea$, we have $g_\patcharea(x, \gamma) = 1$, i.e. the support network is fully activated. For any $x \notin \patcharea$, if for one of $i\in I$, we have $a_ix-b_i \leq -1/\gamma$, then $g_\patcharea(x, \gamma) = 0$.
        
        Observe that $g_\patcharea(x, \gamma)$ cannot be zero if $x$ is very close to $\patcharea$, otherwise 
        the resulting function will be discontinuous and violates the criterion of preservation of CPWL.
        In Theorem~\ref{thm: The Limited Side Effect of Local Patch Neural Networks}, we will prove that we can still 
        guarantee limited side effects on the whole input domain outside of $\patcharea$.
        
            
    \subsection{Affine Patch Functions}\label{sec: affine patch function}
        We consider an affine patch function $p_\patcharea(x) = \boldsymbol{c}x+d$, where matrix $\boldsymbol{c}$ and vector $d$ are undetermined coefficients. 
        In a later section, the design of patch network will ensure that on the patch area $\patcharea$, the repaired network is $f(x)+p_\patcharea(x)$. 
        We will first consider 
        finding appropriate $\boldsymbol{c}$ and $d$ such that
        $f(x)+p_\patcharea(x)$ satisfy the specification on $\patcharea$. 
        
        To satisfy the specification $\Phi$, we need $f(x)+p_\patcharea(x) \in \Phi_{out}$ for all $x\in \patcharea$. To obtain a minimal repair, we minimize $\max_{x\in \patcharea} |p_\patcharea(x)|$.
        Thus, we can formulate the following optimization problem
        \begin{eqnarray}\label{eqn: linear patch function}
            \begin{cases}
                \min_{\boldsymbol{c}, d} \max_{x\in \patcharea} |p_\patcharea(x)| = |\boldsymbol{c}x+d|\\
                (\boldsymbol{c}, d)\in \{(\boldsymbol{c},d)\:|\:f(x)+\boldsymbol{c}x+d\in \Phi_{out}, \forall x\in \patcharea\}
            \end{cases}
        \end{eqnarray}
        
        Notice that this is not a LP since both $\boldsymbol{c}$ and $x$ are variables and we have a $\boldsymbol{c}x$ term in the objective.
        
        In general, one can solve it is by enumerating all the vertices of $\patcharea$. Suppose that $\{v_s| s=1,2,..., S\}$ is the set of vertices of $\patcharea$. Since $\Phi_{out}$ is a convex polytope, we have 
        \begin{eqnarray}
            f(x)+p_\patcharea(x) \in \Phi_{out}\textit{ for all }x \in \patcharea
            \Leftrightarrow f(v_s)+p_\patcharea(v_s) \in \Phi_{out} \textit{ for }s=1,2,..., S.
        \end{eqnarray}
        and 
        \begin{eqnarray}
            & \max_{x\in \patcharea} |\boldsymbol{c}x+d| = \max_{s=1,2,...,S} |cv_s+d|
        \end{eqnarray}
        Hence, we can solve the following equivalent LP.
        \begin{eqnarray}\label{LP: linear patch neural network}
            \begin{cases}
                \min_{\boldsymbol{c}, d} H\\
                H \geq (cv_s+d)_i, H \geq -(cv_s+d)_i, \textit{for } s=1,2,...,S \textit{ and } i =1, 2, ..., m\\
                f(v_s)+p_\patcharea(v_s) \in \Phi_{out} \textit{, for } s=1,2,...,S
            \end{cases}
        \end{eqnarray}
        where $H\in\mathbb{R}$ and will take  $\max_{s=1,2,...,S} |cv_s+d|$ when optimal.
        
    \subsection{Repair via Robust Optimization}\label{sec: repair via robust optimization}
    In general, the number of vertices of a convex polytope can be exponential in the size of its H-representation~\cite{convex-polytope} and enumerating the vertices of a convex polytope is known to be expensive especially when the input dimension is large~\cite{bremner1997complexity}. In this section, we show that we can convert problem (\ref{eqn: linear patch function}) to an LP via \textit{robust optimization}~\cite{ben2009robust} without vertex enumeration and make our algorithm much more efficient. 
    
    The optimization problem (\ref{eqn: linear patch function}) can be converted to the following optimization problem, assuming $\Phi_{out} = \{y|a_{out}\cdot y\leq b_{out}\}$:
    \begin{eqnarray}\label{programming optimization}
        \begin{cases}
            \min_{\boldsymbol{c}, d} H\\
            H \geq H_1
            \textit{, where $H_1$ is the maximum value of the following inner LP }\begin{cases}
                \max_x \boldsymbol{c}x+d\\
                x \in \patcharea
            \end{cases}\\
            H \geq H_2
            \textit{, where $H_2$ is the maximum value of the following inner LP }\begin{cases}
                \max_x -\boldsymbol{c}x-d\\
                x \in \patcharea
            \end{cases}\\
            b_{out} \geq H_3
            \textit{, where $H_3$ is the maximum value of the following inner LP }
            \begin{cases}
                \max_x a_{out}\cdot(f(x)+\boldsymbol{c}x+d)\\
                x \in \patcharea
            \end{cases}\\
        \end{cases}
    \end{eqnarray}
    
    For the inner LPs, since we only care about the maximum value and not the feasible solution that reaches the maximum, we can take the dual of inner LPs to avoid enumerating the vertices of $\patcharea$. Take the first inner LP as an example:
    \begin{eqnarray}
        H_1 = 
        \begin{cases}
            \max_x \boldsymbol{c}x+d\\
            x \in \patcharea
        \end{cases}
        = d+
        \begin{cases}
            \max_x \boldsymbol{c}x\\
            ax\leq b
        \end{cases}
        \stackrel{\text{dual}}{=} d+
        \begin{cases}
            \min_p p'b\\
            a'p=c\\
            p\geq 0
        \end{cases}
    \end{eqnarray}
    
    Therefore, we have the following equivalent LP for optimization problem (\ref{eqn: linear patch function}). Taking the dual of the inner LP to convert the whole problem to an LP is known as taking the \textit{robust counterpart}~\cite{ben2009robust} of the original problem.
    \begin{eqnarray}\label{eqn: programming optimization 2}
        \begin{cases}
            \min_{\boldsymbol{c}, d, p_1, p_2, q, H} H\\
            H \geq p_1'b + d, 
            a'p_1=\boldsymbol{c}, 
            p_1\geq 0\\
            H \geq p_2'b - d, 
            a'p_2=-\boldsymbol{c}, 
            p_2\geq 0\\
            b_{out} \geq q'b + a_{out}(d_f+d), 
            a'q=a_{out}(\boldsymbol{c_f}+\boldsymbol{c}), 
            q\geq 0
        \end{cases}
    \end{eqnarray}
    where $f(x) = \boldsymbol{c_f}x + d_f$.

    \subsection{Single-Region Repairs}
        With a support network $g_\patcharea$ and an affine patch function $p_\patcharea$, we can synthesize the final patch network as follows:
        \begin{eqnarray}\label{eqn: local patch neural network}
            h_\patcharea(x, \gamma) = \sigma(p_\patcharea(x)+K\cdot g_\patcharea(x,\gamma)-K) - \sigma(-p_\patcharea(x)+K\cdot g_\patcharea(x,\gamma)-K)
        \end{eqnarray}
        where $K$ is a vector with every entry is equal to the upper bound of $\{|p_\patcharea(x)|_{+\infty}|x\in X\}$. 
        
        \textbf{Remark:} For $x \in \patcharea$, $g_\patcharea(x, \gamma) = 1$, then we have $h_\patcharea(x, \gamma) = \sigma(p_\patcharea(x))-\sigma(-p_\patcharea(x)) = p_\patcharea(x)$. For $x \notin \patcharea$, $g_\patcharea(x, \gamma)$ goes to zero quickly if $\gamma$ is large. When $g_\patcharea(x, \gamma)=0$, we have $h_\patcharea(x, \gamma) = \sigma(p_\patcharea(x)-K) - \sigma(-p_\patcharea(x)-K) = 0$.
        
        The repaired network $\widehat{f}(x) = f(x) + h_\patcharea(x, \gamma)$. Since $f$ and $h_\patcharea$ are both ReLU DNNs, we have $\widehat{f}$ is also a ReLU DNN.
        We will give the formal guarantees on correctness in Theorem~\ref{thm: soundness}.
        
    \subsection{Multi-Region Repairs}\label{sec: multi-region repair case}
        Suppose there are two linear regions, $\patcharea_1$ and $\patcharea_2$, that need to be repaired, and we have generated the affine patch function $p_{\patcharea_1}(x)$ for $\patcharea_1$ and $p_{\patcharea_2}(x)$ for $\patcharea_2$. 
        
        If $\patcharea_1\cap \patcharea_2 = \emptyset$, then we can repair $f(x)$ with $\widehat{f}(x) = $ $f(x) + h_{\patcharea_1}(x, \gamma) + h_{\patcharea_2}(x, \gamma)$ directly, since $h_{\patcharea_1}(x, \gamma)$ and $h_{\patcharea_2}(x, \gamma)$ will not be nonzero at the same time when $\gamma$ is large enough.
        
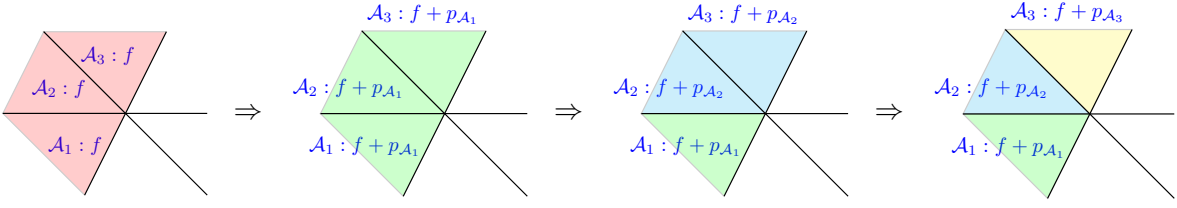
\begin{figure*}
            \centering
            \begin{subfigure}{0.223\textwidth}
                \resizebox{1\textwidth}{!}{
                \begin{tikzpicture}
                    \draw[black,domain=-1.5:1] plot(\x,{0});
                    \draw[black,domain=-1:1] plot(\x,{-\x});
                    \draw[black,domain=-1/2:1/2] plot(\x,{2*\x});
                    \node[left, blue, scale=0.7] at (-0.2, -0.4) {$\patcharea_1: f$};
                    \node[left, blue, scale=0.7] at (-0.4, 0.3) {$\patcharea_2: f$};
                    \node[left, blue, scale=0.7] at (0.2, 0.7) {$\patcharea_3: f$};
                    \node[left, white, scale=0.7] at (0.5, 1.2) {$\patcharea_3: f+p_{\patcharea_1}$};
                    \draw [fill=red, opacity=0.2] (0,0) -- (1/2,1) -- (-1,1) -- cycle;
                    \draw [fill=red, opacity=0.2] (0,0) -- (-1,1) -- (-1.5,0) -- cycle;
                    \draw [fill=red, opacity=0.2] (0,0) -- (-1.5,0) -- (-1/2,-1) -- cycle;
                    \node[left, black] at (1.8, 0) {$\Rightarrow$};
                \end{tikzpicture}}
            \end{subfigure}
            \begin{subfigure}{0.253\textwidth}
                \resizebox{1\textwidth}{!}{
                \begin{tikzpicture}
                    \draw[black,domain=-1.5:1] plot(\x,{0});
                    \draw[black,domain=-1:1] plot(\x,{-\x});
                    \draw[black,domain=-1/2:1/2] plot(\x,{2*\x});
                    \node[left, blue, scale=0.7] at (-0.2, -0.4) {$\patcharea_1: f+p_{\patcharea_1}$};
                    \node[left, blue, scale=0.7] at (-0.4, 0.3) {$\patcharea_2: f+p_{\patcharea_1}$};
                    \node[left, blue, scale=0.7] at (0.5, 1.2) {$\patcharea_3: f+p_{\patcharea_1}$};
                    \draw [fill=green, opacity=0.2] (0,0) -- (1/2,1) -- (-1,1) -- cycle;
                    \draw [fill=green, opacity=0.2] (0,0) -- (-1,1) -- (-1.5,0) -- cycle;
                    \draw [fill=green, opacity=0.2] (0,0) -- (-1.5,0) -- (-1/2,-1) -- cycle;
                    \node[left, black] at (1.8, 0) {$\Rightarrow$};
                \end{tikzpicture}}
            \end{subfigure}
            \begin{subfigure}{0.253\textwidth}
                \resizebox{1\textwidth}{!}{
                \begin{tikzpicture}
                    \draw[black,domain=-1.5:1] plot(\x,{0});
                    \draw[black,domain=-1:1] plot(\x,{-\x});
                    \draw[black,domain=-1/2:1/2] plot(\x,{2*\x});
                    \node[left, blue, scale=0.7] at (-0.2, -0.4) {$\patcharea_1: f+p_{\patcharea_1}$};
                    \node[left, blue, scale=0.7] at (-0.4, 0.3) {$\patcharea_2: f+p_{\patcharea_2}$};
                    \node[left, blue, scale=0.7] at (0.5, 1.2) {$\patcharea_3: f+p_{\patcharea_2}$};
                    \draw [fill=cyan, opacity=0.2] (0,0) -- (1/2,1) -- (-1,1) -- cycle;
                    \draw [fill=cyan, opacity=0.2] (0,0) -- (-1,1) -- (-1.5,0) -- cycle;
                    \draw [fill=green, opacity=0.2] (0,0) -- (-1.5,0) -- (-1/2,-1) -- cycle;
                    \node[left, black] at (1.8, 0) {$\Rightarrow$};
                \end{tikzpicture}}
            \end{subfigure}
            \begin{subfigure}{0.205\textwidth}
                \resizebox{1\textwidth}{!}{
                \begin{tikzpicture}
                    \draw[black,domain=-1.5:1]plot(\x,{0});
                    \draw[black,domain=-1:1] plot(\x,{-\x});
                    \draw[black,domain=-1/2:1/2] plot(\x,{2*\x});
                    \node[left, blue, scale=0.7] at (-0.2, -0.4) {$\patcharea_1: f+p_{\patcharea_1}$};
                    \node[left, blue, scale=0.7] at (-0.4, 0.3) {$\patcharea_2: f+p_{\patcharea_2}$};
                    \node[left, blue, scale=0.7] at (0.5, 1.2) {$\patcharea_3: f+p_{\patcharea_3}$};
                    \draw [fill=yellow, opacity=0.2] (0,0) -- (1/2,1) -- (-1,1) -- cycle;
                    \draw [fill=cyan, opacity=0.2] (0,0) -- (-1,1) -- (-1.5,0) -- cycle;
                    \draw [fill=green, opacity=0.2] (0,0) -- (-1.5,0) -- (-1/2,-1) -- cycle;
                \end{tikzpicture}}
            \end{subfigure}
            \caption{An illustration of multi-region repair with three different repair regions. Left: the original DNN; Middle Left: repair $\patcharea_1\cup\patcharea_2\cup\patcharea_3$ with $p_{\patcharea_1}$; Middle Right: repair $\patcharea_2\cup\patcharea_3$ with $p_{\patcharea_2}-p_{\patcharea_1}$; Right: repair $\patcharea_3$ with $p_{\patcharea_3}-p_{\patcharea_2}$}\label{fig: multi repair}
            \vspace{-2mm}
        \end{figure*}

        However, if $\patcharea_1\cap \patcharea_2 \neq \emptyset$, for any $x\in \patcharea_1\cap \patcharea_2$, both $h_{\patcharea_1}(x, \gamma)$ and $h_{\patcharea_2}(x, \gamma)$ will alter the value of $f$ on $x$, 
        which will invalidate both repairs and cannot guarantee that the repaired DNN will meet the specification $\Phi$. 
        To avoid such over-repairs, 
        our strategy is to first repair $\patcharea_1\cup \patcharea_2$ with $p_{\patcharea_1}(x)$, and then repair $\patcharea_2$ with $p_{\patcharea_2}(x)-p_{\patcharea_1}(x)$. Figure~\ref{fig: multi repair} provides an illustration of a three-region case.
        
        In general, for multi-region repair, we note $\{\patcharea_l\}_{l=1,2,\dots,L}$ are all the buggy linear regions. Then we compute the support network $g_{\patcharea_l}(x, \gamma)$ and affine patch function $p_{\patcharea_l}(x)$ for each $\patcharea_l$. Note that this computation can be done in parallel. 
        
        
        Once we have $g_{\patcharea_l}(x, \gamma)$ and $p_{\patcharea_l}(x)$, we can ``stitch" multiple local patches into a final patch as follows.
        \begin{eqnarray}\label{eqn: multi-region repair}
            h(x,\gamma) =\sum_{l}[\sigma(p_{\patcharea_l}(x) - p_{\patcharea_{l-1}}(x)
            +\max_{j\geq l}\{g_{\patcharea_j}(x, \gamma)\}K_l-K_l)\nonumber\\ - \sigma(-p_{\patcharea_l}(x)+ p_{\patcharea_{l-1}}(x)+\max_{j\geq l}\{g_{\patcharea_j}(x, \gamma)\}K_l-K_l)]
        \end{eqnarray}
        where $K_l$ is the upper bound of $\{|p_{\patcharea_l}(x) - p_{\patcharea_{l-1}}(x)|_{\infty}|x\in X\}$ and $p_{\patcharea_0}(x) = 0$. 
        
        \noindent
        \textbf{Remark}: $\max_{j\geq l}\{g_{\patcharea_j}(x, \gamma)\}$ is a support function for $\cup_{j\geq l} \patcharea_j$ and its value is $1$ for any $x\in \cup_{j\geq l} \patcharea_j$.
        
    \subsection{Feature-Space Repairs}
    \label{sec:feature-space}
        In general, when repairing a large DNN with a high input dimension, the number of linear constraints for one patch area $\patcharea$ will be huge and pose a challenge to solving the resulting LP. One advantage of our approach, which can be used to mitigate this problem, is that it allows for point-wise and area repairs in the feature space in a principled manner, i.e. constructing a patch network starting from an intermediate layer. This approach still preserves soundness and completeness, and is fundamentally different from arbitrarily picking a single layer for repair.
        
        Specifically, for an $R$-layer DNN $f$, we split $f$ into two networks $f_1$ and $f_2$ according to a hidden layer, say the $j_{th}$ hidden layer, where $f_1$ is the function of the first $j$ layers, $f_2$ is the function of the last $R-j$ layers, and $f = f_2\circ f_1$. The output space of $f_1$ is a \textit{feature space}. For any buggy input $\buggypoint$, $f_1(\buggypoint)$ is the corresponding buggy feature. 
        The point-wise repair problem in a feature space is to repair the behavior of $f_2$ on buggy features $\{f_1(\buggypoint_1), \dots, f_1(\buggypoint_L)\}$. Note that this will automatically repair the behavior of $f$ on buggy points $\{\buggypoint_1, \dots, \buggypoint_L\}$.
        
        Repairing in a feature space has the benefit of making the repair process more computation-friendly and reducing the parameter overhead of the additional networks, and has the potential to generalize the repair to undetected buggy inputs with similar features.
        It loses the locality guarantee in the input space but still preserves locality in the feature space.
        
    
\section{Theoretical Guarantees}
    In this section, we present the theoretical guarantees that \toolname provide, and point the readers to proofs of the theorems in the Appendix.
    
    \subsection{Soundness \& Completeness}
        \begin{theorem}[Soundness]\label{thm: soundness}
            The repaired DNN $\widehat{f}$ returned by $\toolname$ is guaranteed to satisfy the specification $\Phi$.
        \end{theorem}

    \begin{theorem}[Completeness]\label{thm: completeness}
            $\toolname$ can always find a solution to the minimal point-wise repair or the minimal area repair problem.
        \end{theorem}

    \subsection{Limited Side Effect of Patch Networks}
        For any $\patcharea$, the support network ensures that the patch network goes to zero quickly when $x$ is away from $\patcharea$. However, it still makes a small change on the neighbors of $\patcharea$. The following theorem shows that for a big enough $\gamma$, the patch network would not change a correct region into incorrect.
        \begin{theorem}[Limited Side Effect]\label{thm: The Limited Side Effect of Local Patch Neural Networks}
            Given a correctness property $\Phi = (\Phi_{in}, \Phi_{out})$, a patch region $\patcharea$ and the corresponding patch network $h(x, \gamma)$, there exists a positive number $\Gamma$ such that for any $\gamma\geq \Gamma$, we have
            \begin{enumerate}
            \item for any linear region $\mathcal{B}\xspace$, if $\mathcal{B}\xspace\cap \patcharea = \emptyset$, then $\widehat{f}(x, \gamma) = f(x)$;
            
            \item for any linear region $\mathcal{C}\xspace$ who is a neighbor of $\patcharea$ ($\mathcal{C}\xspace\cap \patcharea \neq \emptyset$), if $f|_\mathcal{C} \models \Phi$, then $\widehat{f}_\mathcal{C}(x, \gamma) \models \Phi$.
            \end{enumerate}
        \end{theorem}
        
        \begin{corollary}[Incremental Repair]\label{cor}
            For multiple-region repair, the patch for a new region $\patcharea'$ would not cause a previous patched region $\patcharea$ to become incorrect.
        \end{corollary}
        
    \subsection{Minimum Repair}
        \begin{theorem}[Minimum Repair]\label{thm: minimum repair}
            For any ReLU DNN $\tilde{f}$, which is linear on a patch region $\patcharea$ and satisfies the specification $\Phi$, there exists a positive number $\Gamma$, such that for all $\gamma\geq \Gamma$,
            \begin{eqnarray}
                \max_{x\in X}|\tilde{f}(x)-f(x)|\geq \max_{x\in X}|h_\patcharea(x, \gamma)|.
            \end{eqnarray}
        \end{theorem}
        
    \subsection{Efficiency}
        \begin{theorem}[Polynomial-Time Efficiency]
            \toolname terminates in polynomial-time in the size of the neural network and the number of buggy linear regions.
        \end{theorem}
    
    \begin{algorithm}[H]
        \caption{\toolname}
        \label{alg: main}
        \textbf{Input}: A specification $\Phi = (\Phi_{in}, \Phi_{out})$, a ReLU DNN $f$ and a set of buggy points $\{\buggypoint_1,\ldots,\buggypoint_L\} \subset \Phi_{in}$.
        
        \textbf{Output}: A repaired ReLU DNN $\widehat{f}$.
        \begin{algorithmic}[1] 
        \FOR{$l = 1$ to $L$}
        \STATE Generate the patch area $\patcharea_l$ from buggy point $\buggypoint_l$ according to Equation~(\ref{equ: buggy point to linregion});
        \STATE Generate a support network $g_\patcharea$ according to Equation~(\ref{eqn: support neural network});
        \STATE Solve the linear programming problem (\ref{LP: linear patch neural network}) to find the optimal affine patch network $p_\patcharea$.
        \ENDFOR
        \STATE Combine all support networks $g_{\patcharea_l}$ and the corresponding patch networks $p_{\patcharea_l}$ to get the overall patch network $h$ according to Equation~(\ref{eqn: multi-region repair}).
        \STATE \textbf{return} $\widehat{f} = f + h$
        \end{algorithmic}
    \end{algorithm}
\section{Experiments}
    In this Section, we compare \toolname with state-of-the-art methods on both \textit{point-wise repairs} and \textit{area repairs}. The experiments were designed to answer the following questions: 
    \begin{enumerate}
        \item[\textbf{Q1}] \textbf{Effectiveness}: How effective is a repair in removing known buggy behaviors? 
        \item[\textbf{Q2}] \textbf{Locality}: How much side effect (i.e. modification outside the patch area in the function space) does a repair produce?
        \item[\textbf{Q3}] \textbf{Function Change}: How much does a repair change the original neural network in the function space?
        \item[\textbf{Q4}] \textbf{Performance}: Whether and how much does a repair adversely affect the overall performance of the neural network? 
    \end{enumerate}
    
    We consider the following \textbf{evaluation criteria}:
    
        \quad 1. \textit{Efficacy~(E)}: $\%$ of given buggy points or buggy linear regions that are repaired. 
        
        \quad 2. \textit{Norm Difference~(ND)}: average norm ($L_\infty$ or $L_2$) difference between the original DNN and the repaired DNN on a set of inputs (e.g. training and testing data; more details in the tables). We use ND to measure how a repair \textit{change} the original neural network on \textit{function space}. Note that the maximum $L_\infty$ norm difference is $1$ and the maximum $L_2$ norm difference is $\sqrt{2}$.
        
        \quad 3. \textit{Norm Difference on Patch Area~(NDP)}: average norm ($L_\infty$ or $L_2$) difference between the original DNN and the repaired DNN on patch areas (calculated on random sampled points on patch areas or near the buggy points; details in the tables). We use NDP to measure the \textit{locality} of a repair. 
        
        \quad 4. \textit{Accuracy~(Acc)}: accuracy on training or testing data to measure the extent to which a repair preserves the performance of the original neural network.
        
        \quad 5. \textit{Negative Side Effect~(NSE)}: NSE is only for area repair. It is the percentage of correct linear regions (outside of patch area) that become incorrect after a repair. 
        If a repair has a nonzero NSE, the new repair may invalidate a previous repair and lead to a circular repair problem. 
    
        All experiments were run on an Intel Core i5 @ 3.4 GHz with 32 GB of memory. We use Gurobi \cite{gurobi} to solve the linear programs. 
        
        We compared \toolname with the representative related works in Table~\ref{tab: requirements}. 
        \toolname, MDNN and PRDNN guarantee to repair all the buggy points (linear regions). Retrain and Fine-Tuning cannot guarantee 100\% efficacy in general and we run them until all the buggy points are repaired.
    \subsection{Point-wise Repairs: MNIST}
        We train a ReLU DNN on the MNIST dataset~\cite{lecun1998mnist} as the target DNN. 
        It is a multilayer perceptron with ReLU activation functions. It has an input layer with 784 nodes, 2 hidden layers with 256 nodes in each layer, and a final output layer with 10 nodes. 
        
        The goal of a repair is to fix the behaviors of the target DNN on buggy inputs that are found in the test dataset. Thus, the repaired DNN is expected to produce correct predictions for all the buggy inputs. 
        
        The results are shown in Table~\ref{table: pointwise repair}. 
        \toolname achieves almost zero modification outside the patch area (ND) 
        amongst all four methods. In addition, \toolname produces the smallest modification on the patch area (NDP) and preserves the performance of the original DNN (almost no drop on Acc).
        
        \begin{table}
            \centering
            \scalebox{0.93}{
            \begin{tabular}{c ccccc| ccccc}
            & \multicolumn{5}{c}{\toolname} &
            \multicolumn{5}{c}{Retrain \textit{(Requires Training Data)}} \\
            \#P & ND($L_{\infty}$) & ND($L_2$) & NDP($L_{\infty}$) & NDP($L_2$) & Acc & ND($L_{\infty}$) & ND($L_2$)& NDP($L_{\infty}$) & NDP($L_2$) & Acc\\
            \hline
            10 & \textbf{0.0001} & \textbf{0.0002} & \textbf{0.2575} & \textbf{0.3495} & \textbf{98.1\%}   &0.0113 & 0.0155 & 0.7786 & 1.0801 & \textbf{98.1\%}\\
            20 & \textbf{0.0003} & \textbf{0.0003} & \textbf{0.1917} & \textbf{0.2637} & 98.2\%   & 0.0092 & 0.0126 & 0.7714 & 1.0757 & \textbf{98.4\%}\\
            50 & \textbf{0.0006} & \textbf{0.0008} & \textbf{0.2464} & \textbf{0.3355} & 98.5\%   & 0.0084 & 0.0115 & 0.8417 & 1.1584 & \textbf{98.7\%}\\
            100 & \textbf{0.0011} & \textbf{0.0017} & \textbf{0.2540} & \textbf{0.3446} & \textbf{99.0\%}   & 0.0084 & 0.0116 & 0.8483 & 1.1710 & \textbf{99.0\%}\\
            \\
            
            & \multicolumn{5}{c}{Fine-Tuning} &  \multicolumn{5}{c}{PRDNN}\\
            \#P & ND($L_{\infty}$) & ND($L_2$) & NDP($L_{\infty}$) & NDP($L_2$) & Acc & ND($L_{\infty}$) & ND($L_2$)& NDP($L_{\infty}$) & NDP($L_2$) & Acc\\
            \hline
            10 & 0.0220 & 0.0297 & 0.6761 & 0.9238 & 97.6\%   & 0.0141 & 0.0189 & 0.3460 & 0.4736 & 97.8\%\\
            20 & 0.2319 & 0.3151 & 0.8287 & 1.1075 & 78.6\%   & 0.0288 & 0.0387 & 0.4363 & 0.5891 & 97.1\%\\
            50 & 0.3578 & 0.4799 & 0.8473 & 1.1362 & 67.0\%   & 0.0479 & 0.0630 & 0.4937 & 0.6530 & 96.7\%\\
            100 & 0.2383 & 0.3118 & 0.7973 & 1.0796 & 81.9\%   & 0.0916 & 0.1156 & 0.5123 & 0.6534 & 96.1\%\\
            \end{tabular}
            }
            \caption{Point-wise Repairs on MNIST. We use the first hidden layer as the repair layer for PRDNN. The test accuracy of the original DNN is $98.0\%$. \#P: number of buggy points to repair.
            ND($L_\infty$), ND($L_2$): average ($L_\infty$, $L_2$) norm difference on both training and test data. NDP($L_\infty$), NDP($L_2$): average ($L_\infty$, $L_2$) norm difference on random sampled points near the buggy points. Acc: accuracy on test data.
            Note that \toolname automatically performs area repairs on 784-dimensional inputs.
            }
            \label{table: pointwise repair}
            \vspace{-2mm}
        \end{table}
    
        
        We also compare \toolname with MDNN on the watermark removal experiment from their paper. We were not able to run the code provided in the MDNN Github repository, but we were able to run the target DNN models, watermarked images, and MDNN-repaired models from the same repository.
        
        The target DNN is from \cite{goldberger2020minimal}, which has an input layer with 784 nodes, a single hidden layer with 150 nodes, and a final output layer with 10 nodes. The target DNN is watermarked by the method proposed in \cite{adi2018turning} on a set of randomly chosen images $x_i$ with label $f(x_i)$.
        
        The goal is to change the DNN's predictions on all watermarks $x_i$ to any other label $y \neq f(x_i)$ while preserving the DNN's performance on the MNIST test data. 
        For \toolname, we set the prediction $y = f(x_i) - 1$ if $f(x_i) > 1$, and $y = 10$ otherwise.
        
        The results are shown in Table~\ref{tab: watermark}. Both \toolname and MDNN remove all the watermarks. However, MDNN introduces significant distortion to the target DNN and as a result the test accuracy drops rapidly as the number of repair points increases. In comparison, \toolname removes all the watermarks with no harm to test accuracy.
        
        \begin{table}
            \centering
            \scalebox{0.94}{
            \begin{tabular}{c ccccc| ccccc}
            & \multicolumn{5}{c}{\toolname} & \multicolumn{5}{c}{MDNN} \\
            \#P & ND($L_{\infty}$) & ND($L_2$) & NDP($L_{\infty}$) & NDP($L_2$) & Acc & ND($L_{\infty}$) & ND($L_2$)& NDP($L_{\infty}$) & NDP($L_2$) & Acc\\
            \hline
            1 & \textbf{0.000} & \textbf{0.000} & \textbf{0.090} & \textbf{0.128} & $\textbf{96.8\%}$
            & 0.089 & 0.126 & 0.071 & 0.087 & $87.5\%$ \\
            
            5 & \textbf{0.000} & \textbf{0.000} & \textbf{0.127} & \textbf{0.180} & $\textbf{96.8\%}$
            & 0.481 & 0.681 & 0.443 & 0.562 &  $57.1\%$\\
            25 & \textbf{0.000} & \textbf{0.000} & \textbf{0.299} & \textbf{0.420} & $\textbf{96.8\%}$ 
            & 0.904 & 1.278 & 0.637 & 0.815 & $6.7\%$\\
            
            50 & \textbf{0.000} & \textbf{0.000} & \textbf{0.429} & \textbf{0.602} & $\textbf{96.8\%}$ 
            & 0.925 & 1.308 & 0.821 & 1.024 & $4.8\%$\\
            
            100 & \textbf{0.000} & \textbf{0.000} & \textbf{0.462} & \textbf{0.648} & $\textbf{96.8\%}$ 
            & 0.955 & 1.350 & 0.909 & 1.083 & $5.1\%$\\
            \end{tabular}}
            \caption{Watermark Removal. The test accuracy of the original DNN is $96.8\%$. \#P: number of buggy points to repair; ND($L_\infty$), ND($L_2$): average ($L_\infty$, $L_2$) norm difference on both training data and testing data; NDP($L_\infty$), NDP($L_2$): average ($L_\infty$, $L_2$) norm difference on random sampled points near watermark images; Acc: accuracy on test data.}\label{table: watermark removal}
            \label{tab: watermark}
        \end{table}
    \subsection{Area Repairs: HCAS}\label{sec: HCAS}
        To the best of our knowledge, \cite{sotoudeh2021provable} is the only other method that supports area repairs. In this experiment, we compare \toolname with \cite{sotoudeh2021provable} on an experiment where the setting is similar to the \textit{2D Polytope ACAS Xu repair} in their paper.
        
        \cite{sotoudeh2021provable} does not include a vertex enumeration tool (which is required for setting up their LP problem) in their code. We use \texttt{pycddlib} \cite{troffaes2018pycddlib} to perform the vertex enumeration step when evaluating PRDNN. Note that the vertex enumeration tool does not affect the experimental results except running time.
        
        We consider an area repair where the target DNN is the HCAS network (simplified version of ACAS Xu)\footnote{The technique in PRDNN for computing linear regions does not scale beyond two dimensions as stated in their paper. The input space of HCAS is 3D and that of ACAS Xu is 5D so we use HCAS in order to run their tool in our evaluation of area repairs.} $N_{1,4}$ (previous advisory equal to $1$ and time to loss of vertical separation equal to $20s$) from \cite{julian2019guaranteeing}.
        $N_{1,4}$ has an input layer with 3 nodes, 5 hidden layers with 25 nodes in each hidden layer, and a final output layer with 5 nodes. DNN outputs one of five possible control advisories ('Strong left', 'Weak left', 'Clear-of-Conflict', 'Weak right' and 'Strong right').
        
        We use Specification~\ref{spec: HCAS1}, which is similar to Property 5 in \cite{katz2017reluplex}. We use the method from \cite{girard2021disco} to compute all the linear regions for $N_{1,4}$ in the area $\Phi_{in}$ of Specification~\ref{spec: HCAS1} and a total of 87 buggy linear regions were found.
        We apply both $\toolname$ and PRDNN to repair these buggy linear regions. 
        
        \begin{specification}\label{spec: HCAS1}
            If the intruder is near and approaching from the left, the network advises “strong right.” 
            
            Input constraints: $\Phi_{in} = \{ (x, y, \psi) | 10\leq x \leq 5000, 10\leq y \leq 5000, -\pi \leq \psi \leq -1/2\pi\}$. Output constraint: $f(x, y, \psi)_4 \geq f(x, y, \psi)_i$ for $i = 0, 1, 2, 3$.
        \end{specification}
        
        We use Specification~\ref{spec: HCAS2}, the dual of Specification~\ref{spec: HCAS1}, to test the \textit{negative side effect (NSE)} of a repair. We calculate all the linear regions in the area $\Phi_{in}$ of Specification~\ref{spec: HCAS2} and 79 correct linear regions are found. We will test if a repair will make those correct linear regions incorrect.
        
        \begin{specification}\label{spec: HCAS2}
            If the intruder is near and approaching from the right, the network advises “strong left.” 
            
            Input constraints: $\Phi_{in} = \{ (x, y, \psi) | 10\leq x \leq 5000, -5000\leq y \leq -10, 1/2\pi \leq \psi \leq \pi\}$. Output constraint: $f(x, y, \psi)_0 \geq f(x, y, \psi)_i$ for $i = 1, 2, 3, 4$. 
        \end{specification}
        
        The results are shown in Table~\ref{tab: HCAS}. Both \toolname and PRDNN successfully repair all the buggy linear regions. \toolname produces repairs that are significantly better in terms of \textit{locality} (ND), \textit{minimality} (NDP) and \textit{performance preservation} (Acc). 
        In addition, as mentioned in the previous experiment on MNIST, \toolname automatically performs area repair on point-wise repair problems. This means our area repair method scales well to high-dimensional polytopes (the input dimension of MNIST is $784$) whereas PRDNN does not scale beyond 2D linear regions/polytopes.
        
        \begin{table}
            \centering
            \scalebox{0.98}{
            \begin{tabular}{c ccccc| ccccc}
            & \multicolumn{5}{c}{\toolname} & \multicolumn{5}{c}{PRDNN}\\
            \#A & ND($L_{\infty}$) & NDP($L_{\infty}$) & NSE & Acc & T & ND($L_{\infty}$) & NDP($L_{\infty}$) & NSE & Acc & T\\
            \hline
            
            10 & \textbf{0.0000} & \textbf{0.000} & \textbf{0\%} & \textbf{98.1\%} & \textbf{1.0422}
            & 0.0010 & 0.316 & 4\% & 89.6\% & 2.90+0.100\\
            
            20 & \textbf{0.0000} & \textbf{0.022} & \textbf{0\%} & \textbf{98.1\%} & \textbf{1.1856}
            & 0.0015 & 0.372 & 8\% & 83.1\% & 5.81+0.185 \\
            
            50 & \textbf{0.0000} & \textbf{0.176} & \textbf{0\%} & \textbf{98.1\%} & \textbf{1.8174}  
            & 0.0015 & 0.384 & 8\% & 83.8\% & 14.54+0.388\\
            
            87 & \textbf{0.0004} & \textbf{0.459} & \textbf{0\%} & \textbf{97.8\%} & \textbf{2.4571}
            & 0.0014 & 0.466 & \textbf{0\%} & 85.6\% & 25.30+0.714\\
            \end{tabular}
            }
            \caption{Area Repairs on HCAS. 
            We use the the first hidden layer as the repair layer for PRDNN. Results on PRDNN using the last layer (which are inferior to using the first layer) are shown in Table~\ref{apd tab: HCAS} in the Appendix~\ref{sec: Additional Experiment Details}.
            The test accuracy of the original DNN is $97.9\%$. \#A: number of buggy linear regions to repair.
            ND($L_\infty$): average $L_\infty$ norm difference on training data.
            NDP($L_\infty$): average $L_\infty$ norm difference on random sampled data on input constraints of specification~\ref{spec: HCAS1}. 
            NSE: $\%$ of correct linear regions changed to incorrect by the repair. 
            Acc: accuracy on training data (no testing data available). 
            T: running time in seconds. For \toolname, the running time is based on the LP formulation in Appendix~\ref{apd sec: repair via LP}.
            For PRDNN, the first running time is for enumerating all the vertices of the polytopes and the second is for solving the LP problem in PRDNN.}\label{tab: HCAS}
            \vspace{-2mm}
        \end{table}
        
    \subsection{Feature-Space Repairs: ImageNet}
        
        We use AlexNet \cite{krizhevsky2012imagenet} on the ImageNet dataset~\cite{ILSVRC15} as the target DNN. The size of an input image is (224, 224, 3) and the total number of classes for ImageNet is 1000. We slightly modified AlexNet to simplify the evaluation: we only consider 10 output classes that our buggy images may lie on and use a multilayer perceptron with three hidden layers (512, 256, 256 nodes respectively) to mimic the last two layers of AlexNet. 
        The resulting network has 650k neurons, consisting of five convolutional layers, some of which are followed by max-pooling layers, followed by five fully-connected layers with (9216, 4096, 512, 256, 256 nodes respectively).
        The total number of parameters in the resulting network is around 60 million.
        
        The goal of the repair is to fix the behaviors of the target DNN on buggy inputs, which are found on ImageNet-A \cite{hendrycks2021natural}. For $\toolname$, we 
        construct the patch network starting from the 17-th (third from the last) hidden layer (i.e. repair in a feature space). 
        
        The results are shown in Table~\ref{table: pointwise repair ImageNet}. Both \toolname and PRDNN successfully repair all the buggy points.
        \toolname achieves almost zero modification on the validation images compared to the original DNN. In addition, \toolname preserves the performance of the original DNN.
        
        \begin{table}
            \centering
            \scalebox{0.84}{
            \begin{tabular}{c ccc| ccc| ccc| ccc}
            & \multicolumn{3}{c}{\toolname \textit{(feature space)}} &
            \multicolumn{3}{c}{Retrain \textit{(Requires Training Data)}}
            & \multicolumn{3}{c}{Fine-Tuning} 
            &\multicolumn{3}{c}{PRDNN}\\
            \#P & ND($L_{\infty}$) & ND($L_2$) & Acc 
            & ND($L_{\infty}$) & ND($L_2$) & Acc
            & ND($L_{\infty}$) & ND($L_2$) & Acc
            & ND($L_{\infty}$) & ND($L_2$) & Acc\\
            \hline
            
            10 & \textbf{0.0015} & \textbf{0.0018} & \textbf{82.5\%}
            & 0.4343 & 0.5082 & 80.1\%
            & 0.2945 & 0.3563 & 77.9\%
            & 0.2293 & 0.2980 & 82.1\% \\
            
            20 & \textbf{0.0012} & \textbf{0.0015} & 81.3\%
            & 0.4278 & 0.5032 & \textbf{82.9\%}
            & 0.6916 & 0.8103 & 68.5\%
            & 0.2191 & 0.2824 & 80.1\%\\
            
            50 & \textbf{0.0079} & \textbf{0.0099} & 81.3\%
            & 0.5023* & 0.6043* & \textbf{82.1\%}*
            & 0.7669 & 0.8948 & 66.9\%
            & 0.3096 & 0.3799 & 68.9\%
            
            \end{tabular}
            }
            \caption{Point-wise Repairs on ImageNet. PRDNN uses parameters in the last layer for repair. The test accuracy for the original DNN is $83.1\%$. \#P: number of buggy points to repair.
            ND($L_\infty$), ND($L_2$): average ($L_\infty$, $L_2$) norm difference on validation data.
            Acc: accuracy on validation data. * means Retrain only repair 96\% buggy points in 100 epochs.}\label{table: pointwise repair ImageNet}
        \end{table}

\section{Discussion}

\subsection{Parameter Overhead}
$\toolname$ introduces an additional network, patch network, and as a result adds new parameters to the original network. 
The average number of new parameters that $\toolname$ introduces per repair region is in $O(m|I|)$, where $m$ is the dimension of the target DNN's input space and $|I|$ is the number of linear constraints for the H-representation of $\patcharea$.
We can remove redundant constraints in this representation in polynomial time to make the additional network smaller, e.g. iteratively using an LP to check if a constraint is redundant. For the area repair experiment on HCAS, the average number of constraints for one linear region is $3.28$ (which is only $3\%$ of the original $125$ constraints after removing the redundant constraints) and the average number of new parameters that $\toolname$ introduces per repair region is $66$. As a comparison, the number of parameters in the original network is around $3000$ and PRDNN doubles the number of parameters (as a result of the Decoupled DNN construction) regardless of the number of point-wise or area repairs. 
We further note that the removal of redundant constraints can be done offline as a post-repair optimization step. 

Another way to cope with the additional space overhead is to leverage feature-space repairs.
As described in Section~\ref{sec:feature-space}, feature-space repairs allow us to construct the patch network starting from an intermediate layer. 
In addition, for most DNN structures, the dimension of an intermediate layer is typically smaller than the dimension of the input space. 
As a result, the number of additional parameters per repair region will be smaller. 
For the point-wise repair experiment on ImageNet, our feature-space repair adds 500k new parameters on average, which is only $0.1\%$ of the additional parameters introduced if we were to construct the patch network starting from the input layer. 
With feature-space repairs, we are trading-off repair specificity, i.e. how localized the repair is in the input space, with parameter overhead. However, when the number of points or linear regions to repair becomes large, it may make sense to perform repairs in the feature space anyway for better generalization. 
We leave the in-depth investigation of feature-space repairs to future work.

\subsection{Applying \texorpdfstring{$\toolname$}{Lg} To General CPWL Networks}
        Recall the result that an $\mathbb{R}^m \rightarrow \mathbb{R}$ function is representable by a ReLU DNN \textit{if and only if} it is a continuous piecewise linear (CPWL) function~\cite{arora2016relu}. 
        We use convolutional neural networks as an example to show how $\toolname$ can be applied to more general CPWL networks. 
        Convolutional neural networks (CNNs) are neural networks with convolution layers and maxpooling layers. 
        For simplicity, we assume the CNNs also use ReLU activation functions (but in general other CPWL activation functions will also work). 
        The convolutional layers can be viewed as special linear layers. The maxpooling layers can be converted to linear operations with ReLU activation functions as follows.
        \begin{eqnarray*}
            \max(x_1, x_2, ..., x_n) = \max(x_1, \max(x_2, x_3,..., x_n))\\
            \max(x_i, x_j) = \max(x_i-x_j, 0) + x_j = \sigma(x_i-x_j) + x_j
        \end{eqnarray*}
        where $\sigma$ is the ReLU activation function.
        Thus, $\toolname$ can be used to repair CNNs as well.

\section{Conclusion}
We have presented a novel approach for repairing ReLU DNNs with strong theoretical guarantees. Across a set of benchmarks, our approach significantly outperforms existing methods in terms of efficacy, locality, and limiting negative side effects. Future directions include further investigation on feature-space repairs and identifying a lower-bound for $\gamma$.

\bibliographystyle{unsrtnat}
\bibliography{ref}  

\begin{thebibliography}{27}
\providecommand{\natexlab}[1]{#1}
\providecommand{\url}[1]{\texttt{#1}}
\expandafter\ifx\csname urlstyle\endcsname\relax
  \providecommand{\doi}[1]{doi: #1}\else
  \providecommand{\doi}{doi: \begingroup \urlstyle{rm}\Url}\fi

\bibitem[Bojarski et~al.(2016)Bojarski, Testa, Dworakowski, Firner, Flepp,
  Goyal, Jackel, Monfort, Muller, Zhang, Zhang, Zhao, and Zieba]{nvidia-dave}
Mariusz Bojarski, Davide~Del Testa, Daniel Dworakowski, Bernhard Firner, Beat
  Flepp, Prasoon Goyal, Lawrence~D. Jackel, Mathew Monfort, Urs Muller, Jiakai
  Zhang, Xin Zhang, Jake Zhao, and Karol Zieba.
\newblock End to end learning for self-driving cars.
\newblock \emph{CoRR}, abs/1604.07316, 2016.

\bibitem[Shahid et~al.(2019)Shahid, Rappon, and Berta]{nn-healthcare}
Nida Shahid, Tim Rappon, and Whitney Berta.
\newblock Applications of artificial neural networks in health care
  organizational decision-making: A scoping review.
\newblock \emph{PLOS ONE}, 14\penalty0 (2):\penalty0 1--22, 02 2019.
\newblock \doi{10.1371/journal.pone.0212356}.
\newblock URL \url{https://doi.org/10.1371/journal.pone.0212356}.

\bibitem[Julian et~al.(2019)Julian, Kochenderfer, and Owen]{julian2019deep}
Kyle~D Julian, Mykel~J Kochenderfer, and Michael~P Owen.
\newblock Deep neural network compression for aircraft collision avoidance
  systems.
\newblock \emph{Journal of Guidance, Control, and Dynamics}, 42\penalty0
  (3):\penalty0 598--608, 2019.

\bibitem[Dreossi et~al.(2018)Dreossi, Ghosh, Yue, Keutzer,
  Sangiovanni-Vincentelli, and Seshia]{dreossi2018counterexample}
Tommaso Dreossi, Shromona Ghosh, Xiangyu Yue, Kurt Keutzer, Alberto
  Sangiovanni-Vincentelli, and Sanjit~A Seshia.
\newblock Counterexample-guided data augmentation.
\newblock \emph{arXiv preprint arXiv:1805.06962}, 2018.

\bibitem[Ren et~al.(2020)Ren, Yu, Qi, Juefei-Xu, Li, Xue, Ma, and
  Zhao]{ren2020few}
Xuhong Ren, Bing Yu, Hua Qi, Felix Juefei-Xu, Zhuo Li, Wanli Xue, Lei Ma, and
  Jianjun Zhao.
\newblock Few-shot guided mix for dnn repairing.
\newblock In \emph{2020 IEEE International Conference on Software Maintenance
  and Evolution (ICSME)}, pages 717--721. IEEE, 2020.

\bibitem[Sinitsin et~al.(2020)Sinitsin, Plokhotnyuk, Pyrkin, Popov, and
  Babenko]{sinitsin2020editable}
Anton Sinitsin, Vsevolod Plokhotnyuk, Dmitriy Pyrkin, Sergei Popov, and Artem
  Babenko.
\newblock Editable neural networks.
\newblock \emph{arXiv preprint arXiv:2004.00345}, 2020.

\bibitem[Ma et~al.(2018)Ma, Liu, Lee, Zhang, and Grama]{ma2018mode}
Shiqing Ma, Yingqi Liu, Wen-Chuan Lee, Xiangyu Zhang, and Ananth Grama.
\newblock Mode: automated neural network model debugging via state differential
  analysis and input selection.
\newblock In \emph{Proceedings of the 2018 26th ACM Joint Meeting on European
  Software Engineering Conference and Symposium on the Foundations of Software
  Engineering}, pages 175--186, 2018.

\bibitem[Kirkpatrick et~al.(2017)Kirkpatrick, Pascanu, Rabinowitz, Veness,
  Desjardins, Rusu, Milan, Quan, Ramalho, Grabska-Barwinska, Hassabis, Clopath,
  Kumaran, and Hadsell]{catastrophic-forgetting}
James Kirkpatrick, Razvan Pascanu, Neil Rabinowitz, Joel Veness, Guillaume
  Desjardins, Andrei~A. Rusu, Kieran Milan, John Quan, Tiago Ramalho, Agnieszka
  Grabska-Barwinska, Demis Hassabis, Claudia Clopath, Dharshan Kumaran, and
  Raia Hadsell.
\newblock Overcoming catastrophic forgetting in neural networks.
\newblock \emph{Proceedings of the National Academy of Sciences}, 114\penalty0
  (13):\penalty0 3521--3526, 2017.
\newblock ISSN 0027-8424.
\newblock \doi{10.1073/pnas.1611835114}.
\newblock URL \url{https://www.pnas.org/content/114/13/3521}.

\bibitem[Dong et~al.(2020)Dong, Sun, Wang, Wang, and Dai]{dong2020towards}
Guoliang Dong, Jun Sun, Jingyi Wang, Xinyu Wang, and Ting Dai.
\newblock Towards repairing neural networks correctly.
\newblock \emph{arXiv preprint arXiv:2012.01872}, 2020.

\bibitem[Goldberger et~al.(2020)Goldberger, Katz, Adi, and
  Keshet]{goldberger2020minimal}
Ben Goldberger, Guy Katz, Yossi Adi, and Joseph Keshet.
\newblock Minimal modifications of deep neural networks using verification.
\newblock In \emph{LPAR}, volume 2020, page 23rd, 2020.

\bibitem[Sotoudeh and Thakur(2021)]{sotoudeh2021provable}
Matthew Sotoudeh and Aditya~V Thakur.
\newblock Provable repair of deep neural networks.
\newblock In \emph{Proceedings of the 42nd ACM SIGPLAN International Conference
  on Programming Language Design and Implementation}, pages 588--603, 2021.

\bibitem[Arora et~al.(2016)Arora, Basu, Mianjy, and Mukherjee]{arora2016relu}
Raman Arora, Amitabh Basu, Poorya Mianjy, and Anirbit Mukherjee.
\newblock Understanding deep neural networks with rectified linear units.
\newblock \emph{CoRR}, abs/1611.01491, 2016.
\newblock URL \url{http://arxiv.org/abs/1611.01491}.

\bibitem[Serra et~al.(2017)Serra, Tjandraatmadja, and
  Ramalingam]{serra2017linear-region}
Thiago Serra, Christian Tjandraatmadja, and Srikumar Ramalingam.
\newblock Bounding and counting linear regions of deep neural networks.
\newblock \emph{CoRR}, abs/1711.02114, 2017.
\newblock URL \url{http://arxiv.org/abs/1711.02114}.

\bibitem[Lee et~al.(2019)Lee, Alvarez-Melis, and Jaakkola]{lee2019towards}
Guang-He Lee, David Alvarez-Melis, and Tommi~S Jaakkola.
\newblock Towards robust, locally linear deep networks.
\newblock \emph{arXiv preprint arXiv:1907.03207}, 2019.

\bibitem[Henk et~al.(1997)Henk, Richter-Gebert, and Ziegler]{convex-polytope}
Martin Henk, Jürgen Richter-Gebert, and Günter~M. Ziegler.
\newblock Basic properties of convex polytopes.
\newblock In \emph{HANDBOOK OF DISCRETE AND COMPUTATIONAL GEOMETRY, CHAPTER
  13}, pages 243--270. CRC Press, Boca, 1997.

\bibitem[Bremner(1997)]{bremner1997complexity}
David~D Bremner.
\newblock \emph{On the complexity of vertex and facet enumeration for convex
  polytopes}.
\newblock PhD thesis, Citeseer, 1997.

\bibitem[Ben-Tal et~al.(2009)Ben-Tal, El~Ghaoui, and Nemirovski]{ben2009robust}
Aharon Ben-Tal, Laurent El~Ghaoui, and Arkadi Nemirovski.
\newblock \emph{Robust optimization}, volume~28.
\newblock Princeton university press, 2009.

\bibitem[{Gurobi Optimization, LLC}(2021)]{gurobi}
{Gurobi Optimization, LLC}.
\newblock {Gurobi Optimizer Reference Manual}, 2021.
\newblock URL \url{https://www.gurobi.com}.

\bibitem[LeCun(1998)]{lecun1998mnist}
Yann LeCun.
\newblock The mnist database of handwritten digits.
\newblock \emph{http://yann. lecun. com/exdb/mnist/}, 1998.

\bibitem[Adi et~al.(2018)Adi, Baum, Cisse, Pinkas, and Keshet]{adi2018turning}
Yossi Adi, Carsten Baum, Moustapha Cisse, Benny Pinkas, and Joseph Keshet.
\newblock Turning your weakness into a strength: Watermarking deep neural
  networks by backdooring.
\newblock In \emph{27th $\{$USENIX$\}$ Security Symposium ($\{$USENIX$\}$
  Security 18)}, pages 1615--1631, 2018.

\bibitem[Troffaes(2018)]{troffaes2018pycddlib}
Matthias Troffaes.
\newblock pycddlib-a python wrapper for komei fukudals cddlib, 2018.

\bibitem[Julian and Kochenderfer(2019)]{julian2019guaranteeing}
Kyle~D Julian and Mykel~J Kochenderfer.
\newblock Guaranteeing safety for neural network-based aircraft collision
  avoidance systems.
\newblock In \emph{2019 IEEE/AIAA 38th Digital Avionics Systems Conference
  (DASC)}, pages 1--10. IEEE, 2019.

\bibitem[Katz et~al.(2017)Katz, Barrett, Dill, Julian, and
  Kochenderfer]{katz2017reluplex}
Guy Katz, Clark Barrett, David~L Dill, Kyle Julian, and Mykel~J Kochenderfer.
\newblock Reluplex: An efficient smt solver for verifying deep neural networks.
\newblock In \emph{International Conference on Computer Aided Verification},
  pages 97--117. Springer, 2017.

\bibitem[Girard-Satabin et~al.(2021)Girard-Satabin, Varasse, Schoenauer,
  Charpiat, and Chihani]{girard2021disco}
Julien Girard-Satabin, Aymeric Varasse, Marc Schoenauer, Guillaume Charpiat,
  and Zakaria Chihani.
\newblock Disco verification: Division of input space into convex polytopes for
  neural network verification.
\newblock \emph{arXiv preprint arXiv:2105.07776}, 2021.

\bibitem[Krizhevsky et~al.(2012)Krizhevsky, Sutskever, and
  Hinton]{krizhevsky2012imagenet}
Alex Krizhevsky, Ilya Sutskever, and Geoffrey~E Hinton.
\newblock Imagenet classification with deep convolutional neural networks.
\newblock \emph{Advances in neural information processing systems},
  25:\penalty0 1097--1105, 2012.

\bibitem[Russakovsky et~al.(2015)Russakovsky, Deng, Su, Krause, Satheesh, Ma,
  Huang, Karpathy, Khosla, Bernstein, Berg, and Fei-Fei]{ILSVRC15}
Olga Russakovsky, Jia Deng, Hao Su, Jonathan Krause, Sanjeev Satheesh, Sean Ma,
  Zhiheng Huang, Andrej Karpathy, Aditya Khosla, Michael Bernstein,
  Alexander~C. Berg, and Li~Fei-Fei.
\newblock {ImageNet Large Scale Visual Recognition Challenge}.
\newblock \emph{International Journal of Computer Vision (IJCV)}, 115\penalty0
  (3):\penalty0 211--252, 2015.
\newblock \doi{10.1007/s11263-015-0816-y}.

\bibitem[Hendrycks et~al.(2021)Hendrycks, Zhao, Basart, Steinhardt, and
  Song]{hendrycks2021natural}
Dan Hendrycks, Kevin Zhao, Steven Basart, Jacob Steinhardt, and Dawn Song.
\newblock Natural adversarial examples.
\newblock In \emph{Proceedings of the IEEE/CVF Conference on Computer Vision
  and Pattern Recognition}, pages 15262--15271, 2021.

\end{thebibliography}

\section{Appendix}
    \subsection{An Alternative LP Solution}\label{apd sec: repair via LP}
        In Section~\ref{sec: repair via robust optimization}, we show that optimization problem (\ref{eqn: linear patch function}) can be converted to an LP via \textit{robust optimization}~\cite{ben2009robust}. 
        However, taking the dual of the inner LP in robust optimization introduces new variables. Specifically, the number of new variables is in $O(n|I|)$, where $n$ is the DNN's output dimension and $|I|$ is the number of constraints for the linear region. Thus, it is more expensive to solve the LP when $n$ is large (although still much less expensive than enumerating all the vertices). 
        In this section, we show that we can solve optimization problem (\ref{eqn: linear patch function}) via LP more efficiently for many useful cases such as the classification problem in Example~\ref{exp: classification problem} and the HCAS example in Section~\ref{sec: HCAS}.
        
        We consider the case where $\Phi_{out}$ can be expressed as $\{y\:|\: q_l\leq Py\leq q_u\}$ where $P$ is a \textit{full row rank} matrix and $-\infty\leq q_l[i]\leq q_u[i]\leq +\infty$ ($q_l[i]$ and $q_u[i]$ are the $i$-th elements of $q_l$ and $q_u$ respectively).
        
        Consider the following optimization problem.
        \begin{eqnarray}\label{eqn: linear transformation}
            \begin{cases}
            \min_T \max_{x\in \patcharea} |T(f(x) - f(x)|\\
            q_l \leq P(T(f(x)))\leq q_u, \forall x \in \patcharea
            \end{cases}
        \end{eqnarray}
        where $T: \mathbb{R}^n \to \mathbb{R}^n$ is a linear transformation on the DNN's output space $\mathbb{R}^n$. 
        
        \begin{theorem}
            On linear region $\patcharea$, we have $f|_\patcharea(x) = \boldsymbol{f_1} x + f_2$ for some matrix $\boldsymbol{f_1}$ and vector $f_2$. Assuming that $\boldsymbol{f_1}$ is full rank\footnote{
            Note that for neural networks that are trained by a stochastic method, with probability one $\boldsymbol{f_1}$ is full rank.}, the optimization problem in (\ref{eqn: linear patch function}) and the optimization problem in (\ref{eqn: linear transformation}) are equivalent.
        \end{theorem}
        
        Note that $q_l \leq P(T(f(x)))\leq q_u$ can be achieved row by row. Thus, we can find a one-dimensional linear transformation via LPs and combine them into a single linear transformation to solve optimization problem (\ref{eqn: linear transformation}).
        
        For every row of $P$, say the $i$-th row, we can check the lower bound and upper bound of $\{P(f(x))\:|\:\forall x \in \patcharea\}$ on the $i$-th dimension by solving the following LP problems
        \begin{eqnarray}\label{LP: i-th row}
            lb[i] = \min_{x\in \patcharea} P[i](f(x))\quad 
            ub[i] = \max_{x\in \patcharea} P[i](f(x)) 
        \end{eqnarray}
        where $P[i]$ is the $i$-th row of $P$. 
        
        Then for each row $i$, we take a minimal linear transformation $V[i](x) = v_1[i](x) + v_2[i]$ to transfer interval $[lb[i], ub[i]]$ inside interval $[q_l[i], q_u[i]]$. We can take $v_1[i] = 1$ if $q_u[i] - q_l[i] > ub[i] - lb[i]$, else $v_1[i] = \frac{q_u[i] - q_l[i]}{ub[i] - lb[i]}$. And $v_2[i] = q_l[i] - v_1[i] lb[i]$ if $|q_l[i] - v_1[i] lb[i]| \leq |v_1[i] ub[i] - q_u[i]|$, else $v_1[i] ub[i] - q_u[i]$.
        
        Since matrix $P$ is full row rank, we can find a linear transformation $T$ that is equivalent to $V$:
        \begin{eqnarray}
            T = \hat{P}^{-1} V \hat{P} \Rightarrow P(T(f(x))) = V(P(f(x)))
        \end{eqnarray}
        where $\hat{P} = \left[\begin{array}{c}
             P  \\
             P^{\bot} 
        \end{array}\right]$ is an orthogonal extension of $P$ ($P$ and $P^{\bot}$ are orthogonal to each other and $\hat{P}$ is a full rank square matrix).
        
        Once we have $T$, we can obtain an affine patch function $p_\patcharea(x) = T(f(x)) - f(x)$. 
    \setcounter{theorem}{0}
\setcounter{corollary}{0}
\subsection{Proofs of Theorems}
    We prove Theorem \ref{apxthm: soundness} after Corollary \ref{apxcor: Incremental Repair}, since the proof of Theorem \ref{apxthm: soundness} uses the result of Corollary \ref{apxcor: Incremental Repair}.

    \begin{lemma}\label{lemma: soundness for single region}
        The repaired DNN $\widehat{f}$ returned by $\toolname$ is guaranteed to satisfy the specification $\Phi$ on patch area $\patcharea$ in single-region repair case.
    \end{lemma}
    
    \begin{proof}
        By the definition of $p_\patcharea$, we have $f(x)+p_\patcharea(x)\in \Phi_{out}$ for all $x\in \patcharea$.
        
        For any $x \in \patcharea$, we have $g_\patcharea(x, \gamma) = 1$ and $h_\patcharea(x, \gamma) = \sigma(p_\patcharea(x))-\sigma(-p_\patcharea(x)) = p_\patcharea(x)$. Therefore, 
        \begin{eqnarray}
            \widehat{f}(x) = f(x) + h_\patcharea(x, \gamma) = f(x) + p_\patcharea(x) \in \Phi_{out}
        \end{eqnarray}
        
        Thus, the patched neural network $\widehat{f}$ meets the specification $\Phi$ on $\patcharea$.
    \end{proof}
    
    \begin{customthm}{2}[Completeness]\label{apxthm: completeness}
        $\toolname$ can always find a solution to the minimal point-wise repair or the minimal area repair problem.
    \end{customthm}
    
    \begin{proof}
        For every patch area $\patcharea$, we can always find a support network $g_\patcharea$. For any $\Phi_{out}$ and $\patcharea$, there exists an affine function $p_\patcharea$ such that $p_\patcharea(x) \in \Phi_{out}, \forall x\in \patcharea$. Therefore, the LP (\ref{LP: linear patch neural network}) is always feasible and \toolname can find an affine patch function $p_\patcharea$.
        
        Once we have $g_\patcharea$ and $p_\patcharea$ for patch area $\patcharea$, \toolname returns a patch network either by Equation (\ref{eqn: local patch neural network}) or by Equation (\ref{eqn: multi-region repair}).
    \end{proof}
    
    \begin{customthm}{3}[Limited Side Effect]\label{apxthm: Limited Side Effect}
        Given a correctness property $\Phi = (\Phi_{in}, \Phi_{out})$, a patch region $\patcharea$ and the corresponding patch network $h(x, \gamma)$, there exists a positive number $\Gamma$ such that for any $\gamma\geq \Gamma$, we have
        \begin{enumerate}
            \item for any linear region $\mathcal{B}\xspace$, if $\mathcal{B}\xspace\cap \patcharea = \emptyset$, then $\widehat{f}(x, \gamma) = f(x)$;
            \item for any linear region $\mathcal{C}\xspace$ who is a neighbor of $\patcharea$ ($\mathcal{C}\xspace\cap \patcharea \neq \emptyset$), if $f|_\mathcal{C} \models \Phi$, then $\widehat{f}_\mathcal{C}(x, \gamma) \models \Phi$.
        \end{enumerate}
    \end{customthm}
    
    \begin{proof}
        Since a multi-region repair is a composition of multiple singe-region repairs according to Equation~(\ref{eqn: multi-region repair}), we can prove the limited side effect of a multi-region repair by proving the limited side effect of its constituent singe-region repairs. Below, we prove the limited side effect of a singe-region repair.
        
        Consider patch area $\patcharea = \{x\:|\:a_ix\leq b_i, i\in I\}$ and $\patcharea_{> 0}(\gamma) = \{x\:|\:h(x, \gamma) > 0\}$.
    
        1. Since the number of neighbors for $\patcharea$ are finite, we can take a big enough $\gamma$, such that for any $\mathcal{B}\xspace$, if $\mathcal{B}\xspace\cap \patcharea = \emptyset$, $\mathcal{B}\xspace\cap \patcharea_{> 0}(\gamma) = \emptyset$. Thus, we have $\widehat{f}(x, \gamma) = f(x)$ on $\mathcal{B}\xspace$.
        
        2. For any linear region $\mathcal{C}$ who is a neighbor of $\patcharea$, i.e. $\mathcal{C} \neq \patcharea$ and $\mathcal{C} \cap \patcharea \neq \emptyset$, $\widehat{f}$ is no longer a linear function on $\mathcal{C}\xspace$, since there are some hyperplanes introduced by our repair that will divide $\mathcal{C}\xspace$ into multiple linear regions. 
        
        Specifically, those hyperplanes are $\{x\:|\:\gamma(a_ix-b_i) + 1 = 0\}$ for $i\in I$, $\{x\:|\:\sum_{i\in I} g(a_ix-b_i, \gamma)-|I|+1 = 0\}$, $\{x\:|\:p(x)+K\cdot g_\patcharea(x,\gamma)-K = 0\}$ and $\{x\:|\:-p(x)+K\cdot g_\patcharea(x,\gamma)-K = 0\}$.
        
        For any point $x$ in those hyperplanes, it will fall into one of the following four cases.
        \begin{enumerate}[itemindent=1em]
            \item[(a)] $x\in \{x\:|\:\gamma(a_ix-b_i) + 1 = 0\}$ for some $i\in I$, then $g_\patcharea(x, \gamma) = 0$, $h(x, \gamma) = 0$ and $\widehat{f}(x) \in \Phi_{out}$;
            \item[(b)] $x\in \{x\:|\:\sum_{i\in I} g(a_ix-b_i, \gamma)-|I|+1 = 0\}$, then $g_\patcharea(x, \gamma) = 0$, $h(x, \gamma) = 0$ and $\widehat{f}(x) \in \Phi_{out}$;
            \item[(c)] $x\in \{x\:|\:p(x)+K\cdot g_\patcharea(x,\gamma)-K = 0\}$, then $p(x) = K - K\cdot g_\patcharea(x,\gamma)\geq 0$, $-p(x)+K\cdot g_\patcharea(x,\gamma)-K \leq 0$, $h(x, \gamma) = 0$ and $\widehat{f}(x) \in \Phi_{out}$;
            \item[(d)] $x\in \{x\:|\:-p(x)+K\cdot g_\patcharea(x,\gamma)-K\}$, then $p(x) = K\cdot g_\patcharea(x,\gamma)-K\leq 0$, $p(x)+K\cdot g_\patcharea(x,\gamma)-K \leq 0$, $h(x, \gamma) = 0$ and $\widehat{f}(x) \in \Phi_{out}$;
        \end{enumerate}
        
        By the above analysis, we have $\widehat{f}(x) \in \Phi_{out}$ for the boundary of the new linear regions. Since $\widehat{f}$ is linear on the new linear regions and $\Phi_{out}$ is convex, $\widehat{f}(x) \in \Phi_{out}$ for any $x \in \mathcal{C}\xspace$.
    \end{proof}
    
    \noindent
    \textbf{Remark}: By Theorem \ref{apxthm: Limited Side Effect}, we have that a patch would not change a correct linear region to an incorrect one.
    
    \begin{corollary}[Incremental Repair]\label{apxcor: Incremental Repair}
        For multiple-region repair, the patch for a new region $\patcharea'$ would not cause a previous patched region $\patcharea$ to become incorrect.
    \end{corollary}
    
    \begin{proof}
        After applying the patch to linear region $\patcharea$, we have that the resulting network is correct on $\patcharea$. When applying a new patch to another linear region $\patcharea'$, by Theorem \ref{apxthm: Limited Side Effect}, the new patch would not make a correct linear region $\patcharea$ incorrect. 
    \end{proof}

    \begin{customthm}{1}[Soundness]\label{apxthm: soundness}
        The repaired DNN $\widehat{f}$ returned by $\toolname$ is guaranteed to satisfy the specification $\Phi$.
    \end{customthm}
    
    \begin{proof}
        The proof has two parts:
        \begin{enumerate}
            \item to show that $\widehat{f}$ satisfies the specification $\Phi$ on $\patcharea$, and
            \item to show that $\widehat{f}$ satisfies the specification $\Phi$ outside of $\patcharea$.
        \end{enumerate}
        
        Part 1:
        
        Lemma (\ref{lemma: soundness for single region}) shows $\widehat{f}$ satisfy the specification $\Phi$ for single-region repair on $\patcharea$. 
        
        For the multi-region case, consider a set of buggy linear regions $\cup_{1\leq l \leq I}\patcharea_l$ with 
        the corresponding support neural network $g_{\patcharea_l}$ and affine patch function $p_{\patcharea_l}$ for each $\patcharea_l$. For the multi-region repair construction in Equation~(\ref{eqn: multi-region repair}), we refer to $\sigma(p_{\patcharea_j} - p_{\patcharea_{j-1}} +\max_{k\geq j}\{g_{\patcharea_k}\}K_j-K_j)$ as the $j$-th patch and $\widehat{f}_j = f + \sum_{j'\leq j}\sigma(p_{\patcharea_{j'}} - p_{\patcharea_{j'-1}} +\max_{k\geq j'}\{g_{\patcharea_k}\}K_j-K_j)$ as the network after the $j$-th patch.
        
        For any $x$ in patch area $\cup_{1\leq l\leq I} \patcharea_l$, we can find a $j$ such that $x\in \patcharea_j$ but $x\notin \patcharea_{k}$ for all $k>j$. After the first $j$ patches $\sigma(p_{\patcharea_1}(x) +\max_{k\geq 1}\{g_{\patcharea_k}(x, \gamma)\}K_1-K_1)$, $\sigma(p_{\patcharea_2}(x) - p_{\patcharea_1}(x) +\max_{k\geq 2}\{g_{\patcharea_k}(x, \gamma)\}K_2-K_2)$, ... , $\sigma(p_{\patcharea_j}(x) - p_{\patcharea_{j-1}}(x) +\max_{k\geq j}\{g_{\patcharea_k}(x, \gamma)\}K_j-K_j)$, the DNN's output at $x$ becomes $\widehat{f}_j(x) = f(x) + p_{\patcharea_j}(x)$ which meets our specification $\Phi$ at $x$ by the definition of $p_{\patcharea_j}(x)$. 
        
        Since $x\notin \patcharea_{k}$ for all $k>j$, then by Corollary \ref{apxcor: Incremental Repair}, the rest of the patches would not change a correct area to an incorrect area. Therefore, we have the final patched neural network $\widehat{f}$ meets specification $\Phi$ on $\cup_{1\leq l \leq I} \patcharea_l$.
        
        Part 2:
        
        To show that $\widehat{f}$ satisfies $\Phi$ outside of $\patcharea$. 
        
        For any $x$ outside the patch area $\cup_{1\leq l \leq I} \patcharea_l$, we have $x$ lies on a correct linear region (linear region that satisfies the specification $\Phi$). By Theorem \ref{apxthm: Limited Side Effect}, we have either $\widehat{f}(x) = f(x)$ or $\widehat{f}(x) \in \Phi_{out}$. Therefore, $\widehat{f}$ satisfies $\Phi$ outside of $\patcharea$. 
    \end{proof}
    
    \begin{customthm}{4}[Minimum Repair]
        For any ReLU DNN $\tilde{f}$, which is linear on a patch region $\patcharea$ and satisfies the specification $\Phi$, there exists a positive number $\Gamma$, such that for all $\gamma\geq \Gamma$,
        \begin{eqnarray}
            \max_{x\in X}|\tilde{f}(x)-f(x)|\geq \max_{x\in X}|h_\patcharea(x, \gamma)|.
        \end{eqnarray}
    \end{customthm}
    
    \begin{proof}
        We consider the general case where the linear patch function is obtained from Equation~(\ref{eqn: linear patch function}).
        
        For any DNN $\tilde{f}$, which is linear on patch region $\patcharea$ and satisfies the specification $\Phi$, we have $\max_{x\in \patcharea}|\tilde{f}-f|\geq \max_{x\in \patcharea}|cx+d| = \max_{x\in \patcharea}|h_\patcharea(., \gamma)|$ on patch area $\patcharea$ by Equation~(\ref{eqn: linear patch function}).
        
        Therefore, we only need to show:
        \begin{eqnarray}
            \max_{x\notin A}|h_\patcharea(., \gamma)|\leq \max_{x\in A}|h_\patcharea(., \gamma)|
        \end{eqnarray}
        
        Since parameter $\gamma$ controls the slope of $h_\patcharea(., \gamma)$ outside of patch area $\patcharea$, a large $\gamma$ means that $h_\patcharea(., \gamma)$ will drop to zero quickly outside of $\patcharea$. Therefore, we can choose a large enough $\Gamma$ such that $h_\patcharea(., \gamma)$ drops to zero faster than the change of linear patch function $\boldsymbol{c}x+d$. 
        
        Therefore, we have that for any $\gamma\geq \Gamma$,
        \begin{eqnarray}
            \max_{x\notin A}|h_\patcharea(., \gamma)|\leq \max_{x\in A}|h_\patcharea(., \gamma)| \nonumber = \max_{x\in X}|h_\patcharea(., \gamma)| \\
            \leq \max_{x\in A}|\tilde{f}-f|\leq \max_{x\in X}|\tilde{f}-f|
        \end{eqnarray}
        
    \end{proof}
    
    \begin{customthm}{5}[Polynomial-Time Efficiency]
        \toolname terminates in polynomial-time in the size of the neural network and the number of buggy linear regions.
    \end{customthm}
    \begin{proof}
        We consider the affine patch function solved via Robust Optimization~(\ref{eqn: programming optimization 2}).
        Suppose $\patcharea = \{x\in X\:|\:a_i x \leq b_i, i \in I\}$. The running time for solving a Linear Programming is polynomial in the number of variables, and the the number of variables for Linear Programming~(\ref{eqn: programming optimization 2}) is polynomial in $|I|$, which is the number of constraints for $\patcharea$, the DNN's input dimension, and the DNN's output dimension. Since $|I|$ is polynomial in the size of the DNN, \toolname runs in polynomial time in the size of the neural network. 
        
        In addition, since \toolname computes the support network $g_\patcharea$ and affine patch function $p_\patcharea$ for each $\patcharea$ one by one (see Algorithm 1), the time complexity of \toolname is linear in the number of buggy linear regions.
        
    \end{proof}
    
    \begin{customthm}{6}
        On linear region $\patcharea$, we have $f|_\patcharea(x) = \boldsymbol{f_1} x + f_2$ for some matrix $\boldsymbol{f_1}$ and vector $f_2$. Assuming that $\boldsymbol{f_1}$ is full rank\footnote{
        Note that for neural networks that are trained by a stochastic method, with probability one $\boldsymbol{f_1}$ is full rank.}, the optimization problem in (\ref{eqn: linear patch function}) and the optimization problem in (\ref{eqn: linear transformation}) are equivalent.
    \end{customthm}
    \begin{proof}
        On one side, for any $\boldsymbol{c}, d$, since $\boldsymbol{f_1}$ is full rank, there exists a linear transformation $T$, such that $T(f(x)) = T(\boldsymbol{f_1}x+f_2) = (\boldsymbol{f_1}+\boldsymbol{c})x + (f_2+d) = f(x) + \boldsymbol{c}x + d$. 
        
        On the other side, for any $T$, since $T(f(x)) - f(x)$ is linear, there exist $\boldsymbol{c}, d$, such that $\boldsymbol{c}x + d = T(f(x)) - f(x)$.
    \end{proof}
    \subsection{Additional Experiment Details}\label{sec: Additional Experiment Details}
        \textbf{Area Repair: HCAS}
        
        Table~\ref{apd tab: HCAS} is the comparison with PRDNN using the last layer as the repair layer. All other settings are the same as those in Section~\ref{sec: HCAS}.
        
        \begin{table}[H]
            \centering
            \scalebox{0.8}{
            \begin{tabular}{c ccccc| ccccc}
            & \multicolumn{5}{c}{\toolname} & 
            \multicolumn{5}{c}{PRDNN (Last Layer)}\\
            \#A & ND($L_{\infty}$) & NDP($L_{\infty}$) & NSE & Acc & T& ND($L_{\infty}$) & NDP($L_{\infty}$) & NSE & Acc & T\\
            \hline
            10 & \textbf{2.662e-05} & \textbf{1.318e-05} & \textbf{0\%} & \textbf{98.1\%} & \textbf{1.0422}
            & 0.0030 & 0.205 & 16\% & 71.4\% & 2.90+0.100 \\
            
            20 & \textbf{2.918e-05} & \textbf{0.022} & \textbf{0\%} & \textbf{98.1\%} & \textbf{1.1856}
            & 0.0031 & 0.467 & 66\% & 70.5\% & 5.81+0.169 \\
            
            50 & \textbf{8.289e-05} & \textbf{0.176} & \textbf{0\%} & \textbf{98.1\%} & \textbf{1.8174}
            & 0.0031 & 0.467 & 66\% & 70.5\% & 14.54+0.353 \\
            
            87 & \textbf{0.0004} & \textbf{0.459} & \textbf{0\%} & \textbf{97.8\%} & \textbf{2.4571}
            & 0.0031 & 0.467 & 66\% & 70.5\% & 25.30+0.467 \\
            \end{tabular}
            }
            \caption{Area Repairs on HCAS.
            We use the the last hidden layer as the repair layer for PRDNN.
            The test accuracy of the original DNN is $97.9\%$. \#A: number of buggy linear regions to repair; ND($L_\infty$): average $L_\infty$ norm difference on training data ; NDP($L_\infty$): average $L_\infty$ norm difference on random sampled data on input constraints of Specification~\ref{spec: HCAS1}; NSE: $\%$ of correct linear regions that is repaired to incorrect; Acc: accuracy on training data (no testing data available); T: running time in seconds. For PRDNN, the first running time is for enumerating all the vertices of the polytopes and the second is for solving the LP problem in PRDNN.}\label{apd tab: HCAS}
        \end{table}
        
        \textbf{Hyperparameters used in Repair:}
        
        We set learning rate to $10^{-3}$ for Retrain in the point-wise repair experiment.
        
        We set learning rate to $10^{-2}$ and momentum to $0.9$ for Fine-Tuning in the point-wise repair experiment.
        
        PRDNN requires specifying a layer for weight modification. We use the first hidden layer as the repair layer, which has the best performance in our experiment settings, unless otherwise specified.

\end{document}